\documentclass[a4paper,fleqn]{cas-sc}
\usepackage{float,graphicx}
\usepackage{tikz}
\usepackage[numbers]{natbib}

\newcommand{\jl}[1]{\textcolor{black}{#1}}
\begin{document}
\shorttitle{Building age estimation based on geospatial data}
\shortauthors{Jialu Sui et~al.}

\title [mode = title]{Building age estimation based on \jl{widely available multi-source geospatial} data}

\author[1]{Jialu Sui}[style=chinese,
                        orcid=0000-0002-7450-766X]

\author[1,2,3]{Hongsheng Zhang}[style=chinese]
\ead{zhanghs@hku.hk}
\cormark[1]
\author[1]{Rui Liu}[style=chinese]
\author[4]{Jing Ling}[style=chinese]
\author[5]{Ruisheng Wang}[style=chinese]
\author[6]{Hao Wu}[style=chinese]
\author[7,8]{Gang Chen}[style=chinese]

\affiliation[1]{organization={Department of Geography, The University of Hong Kong},
                city={Hong Kong},
                country={China}}

\affiliation[2]{organization={Institute for Climate and Carbon Neutrality, the University of Hong Kong},
                city={Hong Kong},
                country={China}}
\affiliation[3]{organization={The University of Hong Kong Shenzhen Institute of Research and Innovation},
                city={Shenzhen},
                country={China}}
\affiliation[4]{organization={School of Information Engineering, Guangdong University of Technology},
                city={Guangzhou},
                country={China}}

\affiliation[5]{organization={Institute for Smart Cities, School of Architecture and Urban Planning, Shenzhen University},
                city={Shenzhen},
                country={China}}

\affiliation[6]{organization={College of Urban and Environmental Sciences, Academy of Frontier Interdisciplinary Research, Central China Normal University},
                city={Wuhan},
                postcode={430079},
                country={China}}

\affiliation[7]{organization={
Laboratory for Remote Sensing and Environmental Change (LRSEC), Department of Earth, Environmental and Geographical Sciences, University of North Carolina},
                city={Charlotte},
                postcode={NC 28223-0001},
                country={USA}}

\affiliation[8]{organization={Center of Excellence in Regional, Urban, and Built Environmental Analytics (RUBEA), Faculty of Architecture, Chulalongkorn University},
                city={Bangkok},
                postcode={10330},
                country={Thailand}}
                
\cortext[correspondingauthor]{Corresponding Author}


\begin{abstract}
Building age is fundamental for urban infrastructure safety, operation, maintenance, and planning, as it directly affects structural safety, risk management, cost decisions, and sustainable development. Existing methods mostly rely on detailed building information such as street view, function, or materials, which are harder to obtain than \jl{geospatial data} like satellite imagery and Digital Surface Models (DSMs). \jl{However, existing methods based on those geospatial data mainly rely on optical imagery and local texture features, limiting their ability to capture structural information and distinguish buildings from different periods. Moreover, the inherent temporal ordering among building age categories is often overlooked.} To address these limitations, we propose the first deep-learning framework for multi-building age estimation using \jl{widely available multi-source geospatial data, including remote sensing data and their derived products, as well as building footprint information} that estimates age at both pixel and instance levels within a unified pipeline. Specifically, a large pretrained backbone with fine-tuning adapters extracts representative features, and raster-derived and property-derived fusion modules support pixel-wise and instance-wise predictions. Pixel-level estimation employs a pyramid decoder with Cosine-guided Mixture-of-Experts (CMoE) to fuse \jl{multi-source} raster data across scales, capturing fine-grained building age differences, while instance-level prediction uses attention pooling and Property-Conditional Adaptive Instance Normalization (PCAdaIN) to extract instance-level representations and integrate property attributes. A Temporal Ordinal Constraint Loss (TOCL) enforces the sequential ordering of predicted ages. We also construct a building age benchmark for Hong Kong to enable comprehensive evaluation. Experiments achieve RMSE of 9.6919 years and MAE of 6.6978 years. The model is further applied to fill missing building age data across Hong Kong, demonstrating its potential for broader application across diverse cities with low data acquisition requirements. The source code and datasets are available at \href{https://github.com/littlebeen/RS-buildingAge.git}{https://github.com/littlebeen/RS-buildingAge.git}.

\end{abstract}




\maketitle

\section{Introduction}\label{sec:intro}

\begin{figure}
  \centering
   \includegraphics[width=0.5\linewidth]{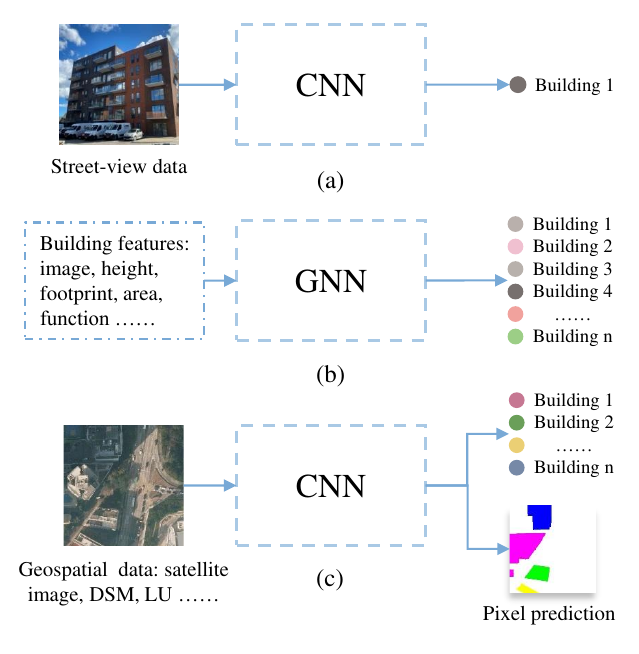}
   \caption{Comparison between our proposed framework and previous deep-learning-based building age estimation methods. Our framework not only provides multi-building instance-level estimation within a unified image-processing pipeline, but also enables pixel-level estimation. Pixel-level and instance-level supervision mutually reinforce each other, leading to improved building age estimation performance. (a) Single-building age estimation based on street-view data. (b) Multi-building age estimation using detailed building features and GNN. (c) Our multi-building age estimation framework based on \jl{widely available multi-source geospatial data}.}
   \label{fig:whole_structure}
\end{figure}

Accurate estimation of building age provides critical information for assessing structural health, identifying potential safety hazards, and guiding maintenance, reinforcement, and renovation strategies \cite{silva2022building}. Such data support informed decision-making in urban renewal \cite{ni2023urban}, disaster prevention \cite{ogawa2023deep}, and public safety management \cite{el2025key}, while also enabling the optimization of maintenance costs and the extension of building service life. With the rapid advancement of remote sensing and artificial intelligence technologies, automatic large-scale building age estimation has emerged as a key approach for enhancing the efficiency of urban surveying and promoting the development of smart cities \cite{paneru2024assessing,aksoezen2015building}.

However, building age information is often incomplete, particularly for structures constructed in the past century, due to gaps in urban architectural records \cite{sun2022understanding}. Compared with other building attributes, such as height or footprint, available building age data are frequently incomplete and inaccurate. \jl{In the existing literature, building age estimation mainly involves two types of tasks, including specific construction-year prediction \cite{CMAB} and predefined age-group classification \cite{nlba,wang2024multi,sun2021automatic}. Existing approaches differ in their spatial prediction scale, ranging from large-scale raster-level estimation \cite{CMAB} to individual-building estimation \cite{tooke2014predicting,sun2021automatic}, with the latter further involving machine-learning \cite{nachtigall2023predicting} and deep-learning paradigms \cite{zeppelzauer2018automatic} based on various types of input data.}
\citet{CMAB} introduced the first multi-attribute building dataset in China, including building age information, but predicted building age at the raster level using long-term impervious surface data. The inherent noise and errors in the impervious surface measurements, coupled with the coarse spatial resolution of the raster data, substantially limit the accuracy of age estimation at the individual building level. Besides, several machine-learning methods \cite{tooke2014predicting,garbasevschi2021spatial,nachtigall2023predicting, rosser2019predicting} have also attempted to estimate building age using basic building attributes. For example, \citet{zhang2025urban} utilized building shape metrics, spatial indicators, and explanatory distance predictors of point-of-interest data to predict building age based on random forest model. \jl{However, these predefined building properties may lack clear and direct correlations with building age, thereby limiting the model’s generalization capability. Regional variations and special building types may distort these original features, thereby reducing prediction accuracy.}

For individual building age estimation, many studies have focused on leveraging detailed building data based on deep-learning methods. As illustrated in Fig.~\ref{fig:whole_structure} (a), several approaches \cite{benz2023towards,sun2021automatic} utilize street-view imagery, which is relatively accessible in most urban areas. For example, \citet{zeppelzauer2018automatic} proposed a two-stage framework for automated building age estimation from unconstrained photographs, which first learns epoch-specific patch-level visual patterns and then aggregates them to obtain instance-level age predictions. \jl{However, these methods face two significant limitations. First, not all buildings are adjacent to streets, resulting in missing street-view images for some structures. Second, since each building requires a separate image for prediction, the feature extraction process for each building image becomes the most time-consuming component of the network, limiting scalability.} Alternatively, as shown in Fig.~\ref{fig:whole_structure} (b), \citet{wang2024multi} and \citet{lu2026progressive} used multi-view graph neural network to establish direct correlations between detailed building attributes and their ages, using property-derived features such as perimeter, area, height, and construction materials. \jl{However, relying solely on explicit correlations derived from a limited set of predefined attributes may constrain predictive performance, since latent age-related cues beyond these manually selected features can be overlooked. In addition, the acquisition of detailed property data for large numbers of buildings is often difficult, making such a strong dependence on property information a major limitation that restricts the applicability and generalization of these methods.}

\jl{Compared with detailed building attribute data, some geospatial data like satellite imagery and Digital Surface Models (DSMs) are more readily accessible across nearly all cities worldwide, thanks to the proliferation of diverse remote sensing sensors. Additionally, these data sources typically cover large geographic areas, making them particularly suitable for studies at the city scale or larger.} \jl{Several methods have utilized semantic segmentation of satellite imagery \cite{CMNeXt, Ftransunet} to predict building attributes like function \cite{du2021mapping,fan2022fine} and material \cite{mantas2025roofsense}.} \jl{This framework can generate age predictions for all buildings within a single image in one forward pass. The model implicitly establishes correlations between 2D visual building features and construction age, avoiding extra noise introduced by explicit handcrafted attribute features.} 
\jl{Compared with street-view imagery and detailed building attributes, data such as satellite imagery and DSM mainly provide overhead observations of buildings, with limited fine-grained visual details and intrinsic building characteristics. Therefore, age-related visual cues are often subtle and difficult to extract from general appearance characteristics.} For instance, \citet{nlba} used satellite imagery to predict the age of multiple buildings within a single image by dividing building masks into different categories, which significantly reduces the processing required for large-scale age estimation. However, this model merely applies existing semantic segmentation models to building age estimation, while differences in task objectives limit the effectiveness of existing semantic segmentation methods for accurate building age prediction. 
\jl{In addition, existing approaches in this category rely solely on optical satellite imagery and lack complementary multi-source information, making it difficult to distinguish visually similar buildings from different periods and to mitigate the interference caused by renovation and reconstruction. Moreover, the original segmentation models mainly capture local texture features while lacking global structural information of buildings, which further limits prediction accuracy. Furthermore, this segmentation-based formulation treats different building age categories independently and does not consider their inherent temporal order, limiting its ability to model the continuous evolution of building characteristics.}

Semantic segmentation methods can generally be categorized into Convolutional Neural Network (CNN)-based models \cite{xu2023lightweight}, Graph Neural Network (GNN)-based models \cite{he2023gnn}, Transformer-based models \cite{strudel2021segmenter}, hybrid models \cite{CMTFNet}, and related variants. More recently, foundation models \cite{Sam,Sam2,DINOv3} have enhanced semantic segmentation performance through their powerful image encoding capabilities, providing rich feature representations that form a solid basis for pixel-level building age estimation. Particularly, DINOv3 \cite{DINOv3} provided rich and diverse large-scale pretrained features, offering strong semantic representations, and several recent studies \cite{mehmood2026remote,yang2025segdino} leveraged it as the model backbone. \jl{However, unlike standard semantic segmentation tasks or other building attribute predictions such as function or material, building age estimation requires the effective integration of both local and global building information, along with fine-grained discrimination capabilities, as differences between buildings constructed in consecutive periods are often subtle. Consequently, task-specific improvements to segmentation models are necessary to capture these subtle, building-specific distinctions.}



\begin{table}[h]
	\centering
	\caption{Comparison of existing open-source building age estimation datasets based on widely available \jl{multi-source geospatial data.}}
	\setlength{\tabcolsep}{1.3mm}{
		\begin{tabular}{cccccc}
			\hline
                \centering\textbf{Dataset}&\textbf{Acquired Place}&\textbf{Image Number}&\textbf{Building Number}&\textbf{Resolution (m)}&\textbf{Input Data Type}\\
                \hline
                MapYourCity \cite{MapYourCity}& European cities&30,849&30,849&0.5&Street View, Satellite\\
                NLBA \cite{nlba}& Netherlands &5,000&Unknown&0.25& Satellite, \jl{Footprint}\\
                \hline
                Hong Kong&Hong Kong&4,194&\jl{29,958}&0.25& Satellite, DSM, LU, Footprint \\
		\hline
	\end{tabular}}
	\label{tab:datasets}
\end{table}

Moreover, the scarcity of suitable datasets further limits the development of building age estimation methods. There are relatively few open-source datasets available for segmentation-based building age estimation, and existing datasets exhibit notable limitations. As summarized in Table~\ref{tab:datasets}, \citet{MapYourCity} introduced a building age estimation dataset based on satellite and street-view imagery. However, each data sample corresponds to only a single building, resulting in age annotations for 30,849 buildings, which is equal to the number of images. In contrast, \citet{nlba} provided pixel-level age classification results for satellite imagery, but does not include explicit building instance annotations, leaving the exact number of buildings unspecified. These datasets either focus on isolated building samples or lack explicit instance-level annotations, limiting their suitability for unified pixel-level and instance-level building age estimation.

To address these challenges, \jl{we propose a segmentation-based deep-learning model to explore the intrinsic relationships between building features and building age.} To the best of our knowledge, this is the first work to perform building age estimation at both pixel and instance levels within a unified image-processing pipeline using \jl{widely available multi-source geospatial data, including remote sensing data and their derived products, as well as building footprint information.} The generation of pixel-level results enables age estimation without individual footprints, and pixel-level features and outputs, constrained by pixel-level loss, provide auxiliary guidance for instance-level result generation. The framework employs a large pretrained model with a fine-tuning adapter as its backbone for feature extraction. To support both pixel-level and instance-level estimation tasks, we design two complementary fusion branches, a raster-derived data fusion module and a property-derived data fusion module. For pixel-level estimation, a pyramid decoder with Cosine-guided Mixture-of-Experts (CMoE) integrates raster-derived features across scales to capture fine-grained building age differences. For instance-level estimation, attention pooling and Property-Conditional Adaptive Instance Normalization (PCAdaIN) extract instance-level representations and fuse building features with property attributes, focusing on intrinsic characteristics while reducing surrounding interference. A Temporal Ordinal Constraint Loss (TOCL) is applied during model training to preserve the temporal hierarchy of the estimated building ages. Furthermore, we construct a building dataset based on \jl{widely available multi-source geospatial data} from Hong Kong to facilitate model evaluation and support research on building age estimation.

In summary, the main contributions of this work are as follows:
\begin{enumerate}
\item We propose the first framework for multi-building age estimation based on widely available \jl{multi-source geospatial data}, which enables both pixel-level and instance-level estimation within a single, integrated processing pipeline.


\item We propose a multi-branch feature fusion framework built on a large pretrained model. A pyramid decoder with CMoE handles raster-derived fusion and pixel-level estimation to capture fine-grained age variations, while attention pooling and PCAdaIN enable instance-level learning with property fusion. In addition, TOCL is used to enforce the ordinal consistency of building age predictions.

\item To evaluate the performance of our model, we gather existing building data from Hong Kong to construct a \jl{multi-source geospatial} dataset for building age estimation. Furthermore, after training on this data, we apply our model to the whole of Hong Kong to fill in the missing building age data and use the development history of Hong Kong to validate our model’s performance.
\end{enumerate}

\begin{figure}
  \centering
   \includegraphics[width=\linewidth]{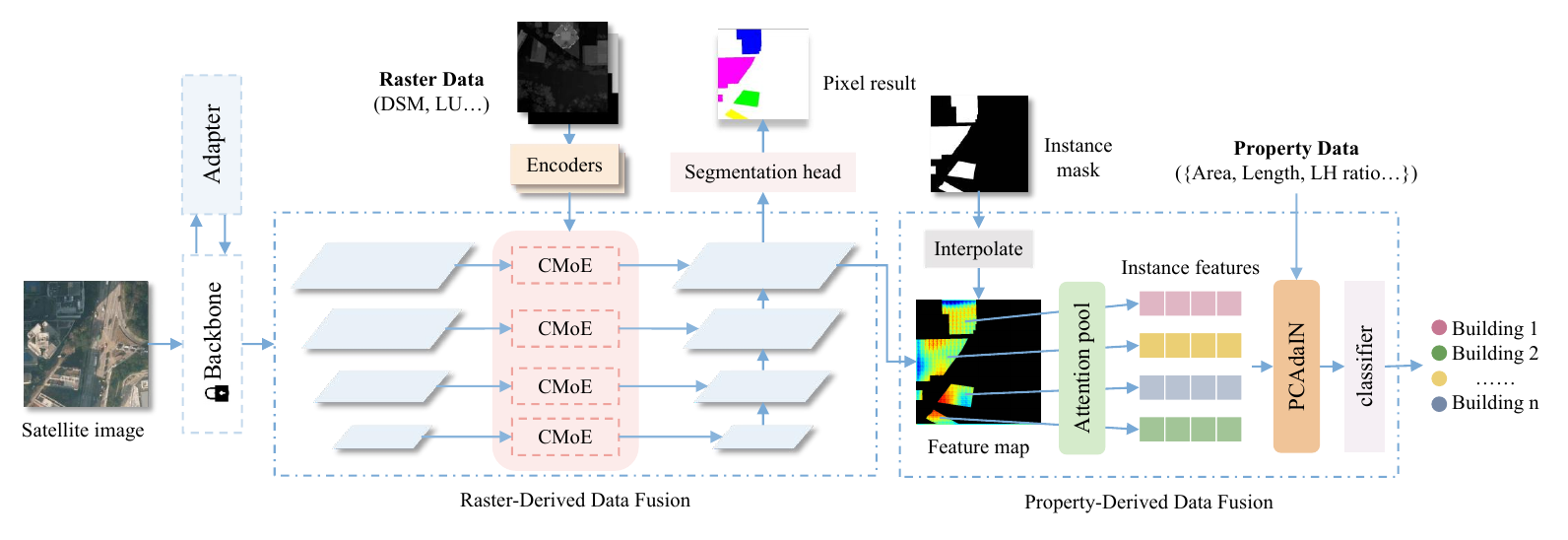}
   \caption{The proposed framework consists of two complementary fusion branches: a raster-derived data fusion module for pixel-level estimation and a property-derived data fusion module for instance-level estimation.}
   \label{fig:framework}
\end{figure}

The remainder of this paper is structured as follows: Section~\ref{sec:method} and Section~\ref{sec:datasets} introduce the proposed network and dataset. After that, Section~\ref{sec:experiments} presents the experimental findings, whereas discussion and insights are deliberated in Section~\ref{sec:discussion}. Finally, concluding remarks are provided in Section~\ref{sec:conclusion}.

\section{Method}\label{sec:method}

\subsection{Problem statement}
\jl{To tackle the fine-grained classification of building age, we propose the first framework for multi-building age estimation based on widely available multi-source geospatial data, including remote sensing data and their derived products, as well as building footprint information. The proposed framework incorporates instance-level building representations to capture structural characteristics and enables both pixel-level and instance-level age estimation within a unified processing pipeline. It takes satellite imagery and building footprints as essential inputs, while flexibly integrating auxiliary multi-source data according to their availability. The auxiliary data can be categorized into raster-derived data and property-derived data. This flexible design enhances the scalability and adaptability of the framework across urban regions with different levels of data availability. By taking the entire satellite image as input, the model can exploit surrounding environmental contexts beyond individual building appearances. Meanwhile, auxiliary data provide additional age-related cues that are less affected by visual similarities among buildings from different periods and appearance changes caused by renovation or reconstruction, enabling more reliable building age estimation.}

\subsection{Framework}
This framework performs multi-building age estimation within a unified image-processing pipeline, as illustrated in Fig.~\ref{fig:framework}. \jl{Specifically, the satellite image is first fed into the network as the fundamental condition, after which a pretrained vision foundation model, DINOv3 \cite{DINOv3}, is employed for high-level feature extraction and generates robust visual representations with a focus on capturing age-sensitive features such as rooftop texture degradation, era-characteristic architectural details, material weathering patterns, and spatial context.} \jl{To facilitate effective task-specific adaptation while preserving the pretrained knowledge of the foundation model, an adapter module \cite{MFNet} is incorporated into the backbone architecture, with only the adapter parameters being updated during training, thereby ensuring parameter-efficient fine-tuning.}

Upon completion of the foundational feature extraction stage, two complementary fusion mechanisms, namely raster-derived data fusion and property-derived data fusion, are introduced to support building age estimation at both the pixel and instance levels. Pixel-level prediction allows building age estimation without individual footprints, while pixel-level features and outputs supervised by pixel-level loss provide supportive constraints for instance-level prediction. In the raster-derived data fusion stage, auxiliary geospatial raster data, such as DSM and Land Utilization (LU) data, are integrated with the RGB-based satellite imagery, enabling the model to jointly exploit optical, structural, and contextual information. After the \jl{multi-source data} fusion process, the fused features are passed through a segmentation head to generate pixel-level building age estimation, thereby providing spatially detailed building age estimation across the entire scene. Subsequently, individual building instance features are extracted using instance masks from the raster feature representations. Building-related property attributes, such as building area, building length and the Length-Height (LH) ratio, are then incorporated and fused with the corresponding instance features. Finally, by jointly integrating the fused raster features and structural property attributes, the framework performs instance-level age estimation for each building present in the satellite image.

\begin{figure}
  \centering
   \includegraphics[width=0.7\linewidth]{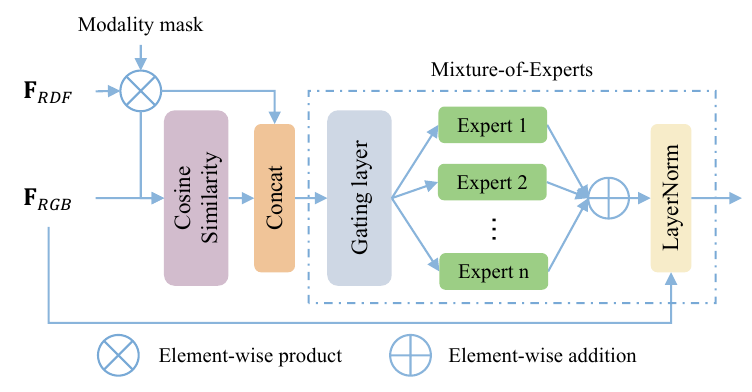}
   \caption{The structure of CMoE, which contains three modules: a modality mask gate, a cosine similarity adaptation module, and an MoE module.}
   \label{fig:moe}
\end{figure}

\begin{figure}
  \centering
   \includegraphics[width=0.6\linewidth]{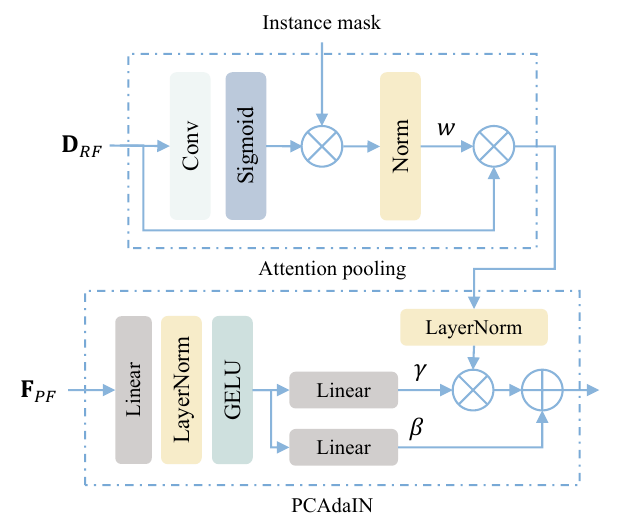}
   \caption{The structures of Attention Pooling and PCAdaIN. Attention Pooling extracts instance-level features from the whole-image feature representations, while PCAdaIN fuses building property data with the corresponding raster features.}
   \label{fig:AdaIN}
\end{figure}

\subsection{Raster-Derived Data Fusion for Pixel-Level Estimation}

All auxiliary raster data are first processed by dedicated encoders to obtain multi-scale feature representations. In the raster-derived data fusion stage, the features extracted by the foundation model are further projected to multiple scales, enabling effective integration with the auxiliary raster-derived features within a pyramid-structured fusion framework. Through this hierarchical fusion process, implemented with a set of CMoE modules operating at multiple scales, \jl{multi-source data} features from different resolutions are effectively aligned and aggregated, enabling the capture of fine-grained building age variations for precise age classification. Subsequently, the fused multi-scale features are processed by a decoder \cite{UNetFormer}, which progressively reconstructs spatial representations while preserving semantic information. Finally, a segmentation head is added to perform pixel-level prediction by generating the probability distribution over age classes for each building pixel.

\subsubsection{CMoE}

We propose the CMoE module to enable adaptive \jl{multi-source data} feature integration. The module consists of three components: a modality mask gate, a cosine similarity adaptation module, and a Mixture-of-Experts (MoE) module. \jl{As illustrated in Fig.~\ref{fig:moe}, firstly, a modality mask $\mathcal{M} = [m_1, m_2, \dots, m_J]^\top,\quad m_j \in \{0, 1\}$ is introduced to dynamically regulate the availability of each input modality, ensuring that the contribution of a modality is suppressed whenever it is absent. Subsequently, the cosine similarity with the hyperparameter $\varepsilon$ between the features from auxiliary modalities $\mathbf{F}_{RDF}=\{\mathbf{F}_{DSM},\mathbf{F}_{LU}, \dots \}$ and the primary image modality $\mathbf{F}_{RGB}$ is computed to guide feature alignment, which improves robustness under varying modality configurations and enables a plug-and-play extension mechanism for additional modalities.} The detailed formulation is given as follows:
\begin{equation*}
	\omega_{cos} = (1+ \varepsilon * \text{cos}(\mathbf{F}_{RDF} \odot \mathcal{M}, \mathbf{F}_{RGB})),
\end{equation*}
\begin{equation*}
	\mathbf{F}_{cos}= \text{Concat}(\omega_{cos} \odot \mathbf{F}_{RGB},\mathbf{F}_{RDF}),
\end{equation*}
where $\text{cos}(\cdot)$ and $\text{Concat}(\cdot)$ denote the cosine similarity operation and matrix concatenation, respectively.

\jl{Finally, the MoE gating mechanism assigns different experts to handle diverse modality combinations and scene conditions, allowing the expert weights to vary spatially and adaptively according to the input content, thereby enhancing the model’s ability to capture diverse and fine-grained age-related patterns.} Specifically, a convolutional gating layer processes the auxiliary raster-derived features to generate spatially aware expert weights $\mathbf{G}$, which are subsequently normalized using a softmax function. Each expert $\text{Expert}_i(\cdot)$ is implemented as a convolutional layer that learns specialized representations from $F_{cos}$. The final fused features are then computed as the weighted sum of all expert outputs based on the gating weights, and further refined through a residual connection followed by layer normalization. The final fused features $F_{RF}$ can be modeled as:
\begin{equation*}
	\mathbf{G} = \text{Softmax}\big(\text{Conv}(\text{Concat}(\mathbf{F}_{RGB}, \mathbf{F}_{RDF} \odot \mathcal{M}))\big),
\end{equation*}
\begin{equation*}
\mathbf{F}_{RF} = \text{LayerNorm}\big(\mathbf{F}_{RGB} + \sum_{i=1}^S \mathbf{G}_i \odot \text{Expert}_i(\mathbf{F}_{cos})\big),
\end{equation*}
where $S$ is the number of experts.

\subsection{Property-Derived Data Fusion for Instance-Level Estimation}

Property-derived data fusion is designed to generate robust instance-level representations that integrate both spatial and property information for accurate building age estimation. Initially, instance features are extracted based on the instance masks, and attention pooling is applied to aggregate the corresponding pixel-level features into a global representation for each instance. After obtaining the instance features, the building property attributes are fused with the corresponding instance features derived from the raster data. A classifier is finally employed to perform the final age estimation.

\subsubsection{Attention pooling}
\jl{As shown in the upper part of Fig.~\ref{fig:AdaIN}, the attention pooling module leverages the instance mask $\mathbf{M}$ to restrict feature aggregation of feature $\mathbf{D}_{RF}$ after the decoder within valid regions, automatically learning spatial attention weights to highlight informative areas while suppressing background noise.} Specifically, $\mathbf{D}_{RF}$ is first passed through a convolutional layer to generate a spatial attention map, which is activated by a sigmoid function and multiplied element-wise with the input mask to obtain masked attention weights. These weights are subsequently normalized over the spatial dimensions. The input features are then reweighted using the normalized attention map and aggregated by summing across all spatial locations, yielding channel-wise global pooled features. By performing weighted sum pooling exclusively within the masked regions, the module produces a clean and background-free global representation for each building instance, effectively reducing feature contamination while preserving discriminative structural details. The final output feature $\mathbf{F}_{IF}$ is:
\begin{equation*}
	\mathbf{F}_{IF}=\text{Norm}(\text{Sigmoid}(\text{Conv}(\mathbf{D}_{RF})) \odot \mathbf{M}_j) \odot \mathbf{D}_{RF}.
\end{equation*}

\subsubsection{PCAdaIN}
\jl{In the lower part of Fig.~\ref{fig:AdaIN}, PCAdaIN is employed to achieve smooth and effective multi-source data fusion while preserving the detailed structural information inherent in instance features. Pixel-level instance features extracted from multi-source raster data contain high-dimensional spatial information, whereas property features, such as building area and perimeter, are represented as compact low-dimensional attributes. The substantial differences in feature dimensionality and representation forms make their direct fusion challenging. PCAdaIN addresses this issue by operating in the statistical distribution space, which enables the compact property information to globally regulate the statistical distribution of high-dimensional instance features.}
The input $\mathbf{F}_{PD} = \{Area, Length, LH\ ratio, Function, \dots \}$ represents the set of property attributes corresponding to a given building instance. These property features are first transformed via a feature projection module, consisting of a linear layer, layer normalization, and GELU activation, thereby mitigating the dimensional discrepancy between the low-dimensional property data and the high-dimensional instance features $\mathbf{F}_{IF}$. Following this projection, an Adaptive Instance Normalization (AdaIN)-based mechanism is employed to modulate the instance features according to the projected property features, thereby enabling coherent and semantically aligned \jl{multi-source data} fusion. Specifically, the projected property features are further transformed through linear layers to generate channel-wise scaling $\gamma$ and shifting $\beta$ parameters with the same dimensionality as the instance features. The normalized instance features are then adaptively modulated using these geometry-aware parameters, producing a fused feature representation that effectively incorporates structural property information. The detailed formulation is provided as follows:
\begin{equation*}
	\gamma=\text{Linear}_1(\text{Project}(\mathbf{F}_{PD})),
\end{equation*}
\begin{equation*}
	\beta=\text{Linear}_2(\text{Project}(\mathbf{F}_{PD})),
\end{equation*}
\begin{equation*}
	\mathbf{P}=\text{LayerNorm}(\mathbf{F}_{IF}) \odot \gamma+\beta .
\end{equation*}

\subsection{Loss Function}

Two types of loss functions, standard cross-entropy loss and Temporal Ordinal Constraint Loss (TOCL), are employed to supervise pixel-level and instance-level estimations, respectively. For pixel-level estimation, the standard cross-entropy loss treats each pixel as an independent classification target, which effectively captures fine-grained visual cues such as texture, roof condition, and material while stabilizing training and producing dense, high-quality prediction maps. \jl{The loss calculation ignores background pixels and is computed solely over labeled building areas.} The formulation is given as follows:
\begin{equation*}
	\mathcal{L}_{\text{CE}} = -\frac{1}{N}\sum_{i=1}^N \sum_{k=1}^K y_{i,k} \log z_{i,k},
\end{equation*}
where $K$ is the number of classes and $N$ is the number of samples. $y$ and $z$ represent the ground-truth label and predicted probability.

\jl{For instance-level supervision, we employ TOCL, which optimizes the binary cross-entropy between predicted logits and ordinal targets. Since each pixel only reflects local visual patterns, ordinal constraints are unnecessary at the pixel level, whereas instance-level estimation requires modeling the natural sequential order of building age. TOCL imposes stronger penalties on misclassifications involving larger temporal distances and encourages chronologically consistent predictions, thereby reducing large deviations from the ground truth. In addition, it decomposes the task into multiple binary ordinal subtasks, improving robustness to noise.} The detailed formulation of TOCL is given as follows:
\begin{equation*}
\mathcal{L}_{\text{ordinal}} = \frac{1}{N \cdot K} \sum_{i=1}^N \sum_{k=1}^K \text{BCE}\left( z_{i,k},\; y_{i,k}^{\text{ord}} \right),
\end{equation*}
\begin{equation*}
\text{BCE}(z, y) = - \left[ y \cdot \log(\sigma(z)) + (1-y) \cdot \log(1-\sigma(z)) \right],
\end{equation*}
\begin{equation*}
y_{i,k}^{\text{ord}} =
\begin{cases}
1, & \text{if } k < y_i, \\
0, & \text{otherwise},
\end{cases}
\end{equation*}
where $\sigma(\cdot)$ denotes the sigmoid function.

\begin{table}[tbp]
	\centering
	\caption{Data source of the Hong Kong dataset}
	\setlength{\tabcolsep}{1.3mm}{
		\begin{tabular}{ccc}
			\hline
                \centering\textbf{Data type}&\textbf{Acquisition time}&\textbf{Resolution (m)}\\
                \hline
                True Digital Orthophoto &2025&0.25\\
                DSM&2021&0.5\\
                LU &1990, 2000, 2010, 2020&10 \\
                Building footprint &2025&Vector data \\
                Building age &2026&Vector data \\
		\hline
	\end{tabular}}
	\label{tab:datasetsdetail}
\end{table}

\section{Hong Kong Dataset} \label{sec:datasets}
Hong Kong has developed into one of the most densely built and highly organized urban environments in the world, where the highly complex distribution of building age poses significant challenges for accurate age estimation. \jl{To address this, we construct a Hong Kong dataset by integrating widely available multi-source geospatial data, including satellite imagery, DSM, building footprint data and Land Utilization (LU). LU refers to the usage patterns of land for different purposes, and in this study, it is primarily distinguished by whether the land is used for buildings or not. Detailed dataset statistics and comparisons with existing datasets are provided in Table~\ref{tab:datasets}.} \jl{Notably, although our dataset contains only 4,194 labeled images, it includes annotations for 29,958 building instances. By using only one-sixth of the number of images in \cite{MapYourCity}, our dataset still provides approximately the same number of annotated buildings. The number of buildings is calculated by summing the number of buildings contained in each labeled image.} \jl{Existing building age records are used to annotate a subset of buildings for model training and evaluation. Detailed information about the source data is provided in Table~\ref{tab:datasetsdetail}. Except for LU \cite{xiong202640}, all the other data are obtained from \href{https://portal.csdi.gov.hk/geoportal}{https://portal.csdi.gov.hk/geoportal}.} Subsequently, the trained model is applied to estimate age information for all 315,337 buildings across Hong Kong.


\begin{figure}
  \centering
   \includegraphics[width=0.5\linewidth]{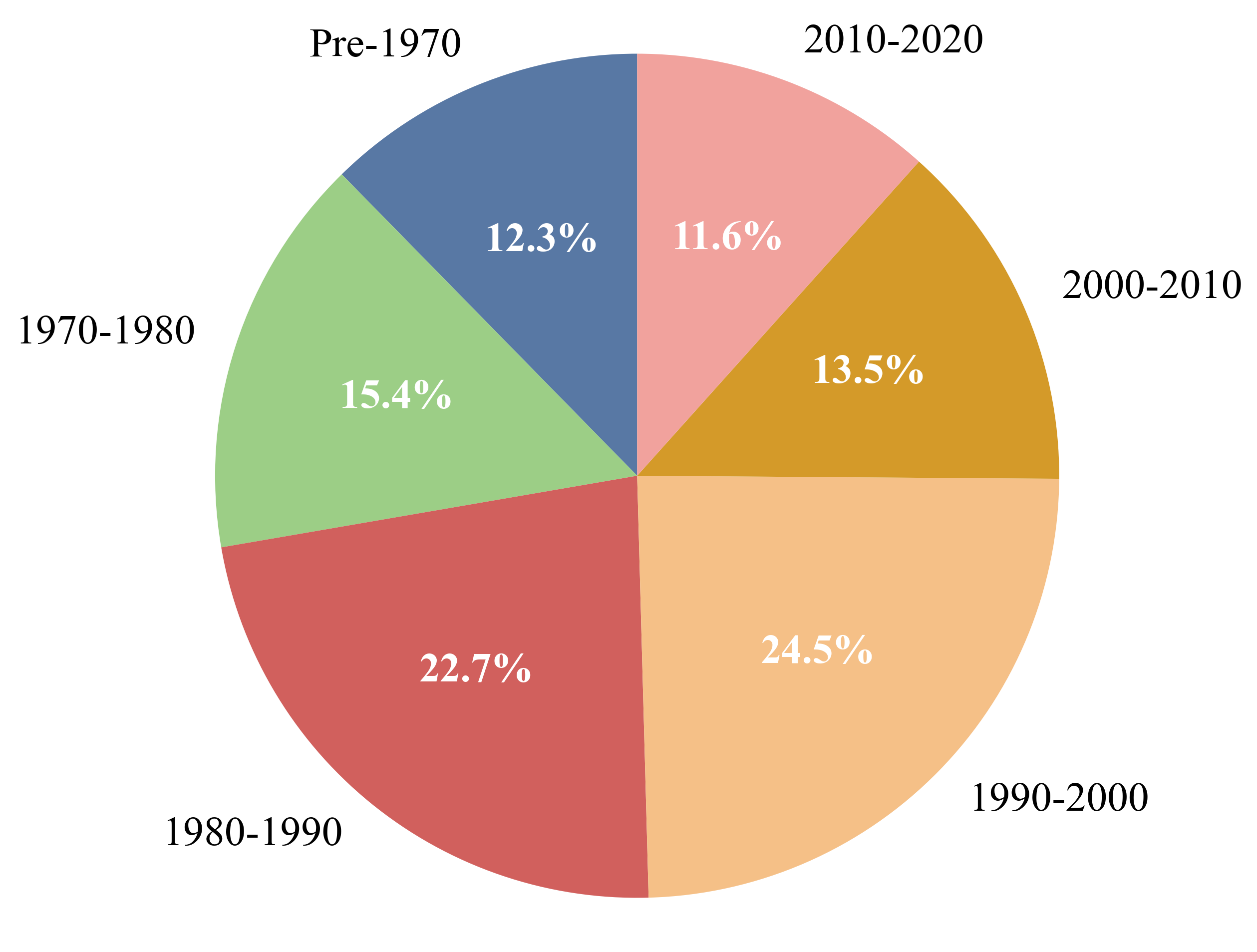}
   \caption{The distribution of building pixels across different construction periods in the labeled dataset.}
   \label{fig:6classes}
\end{figure}

\begin{figure}
  \centering
   \includegraphics[width=0.6\linewidth]{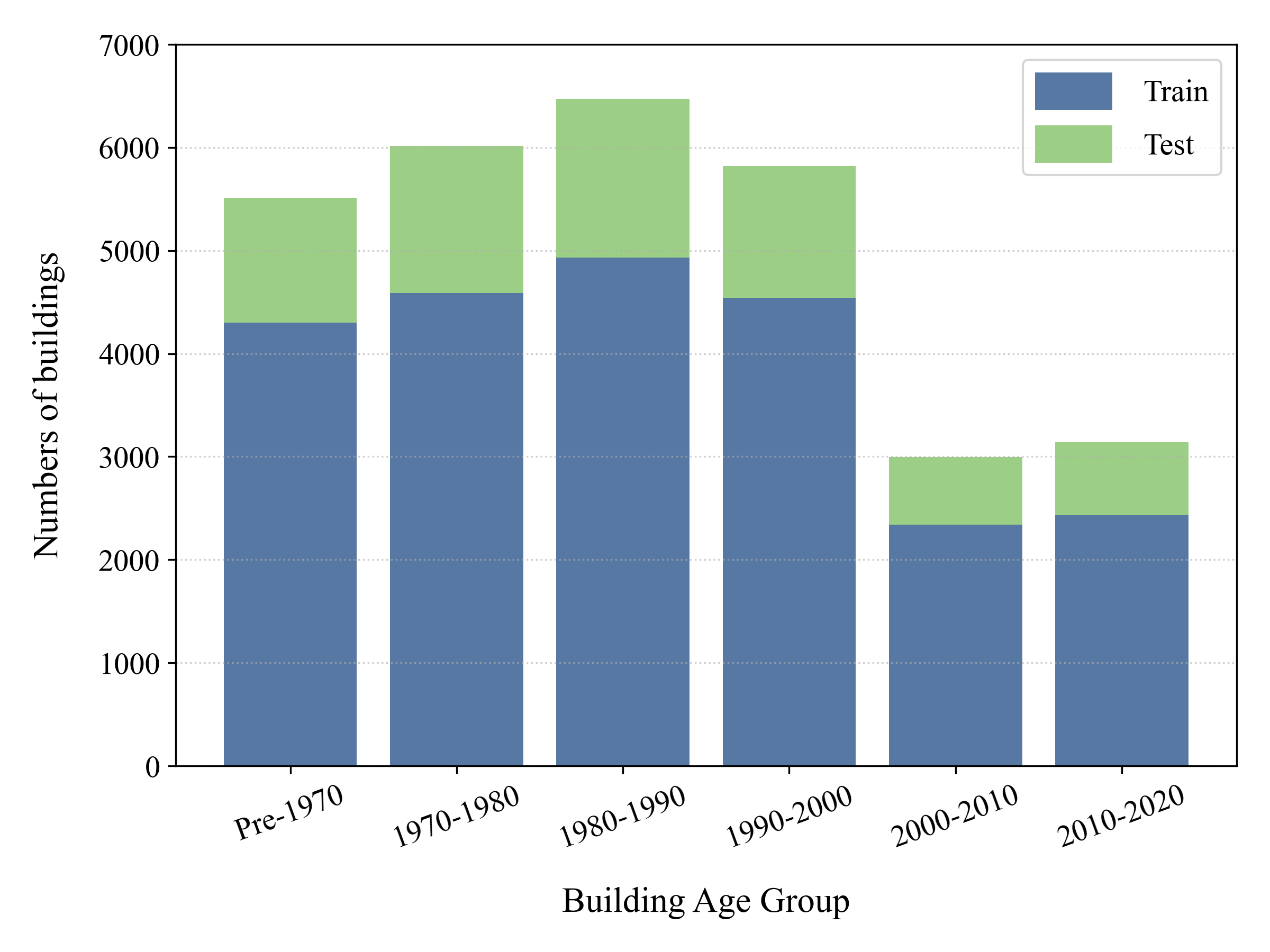}
   \caption{\jl{The distribution of buildings across different construction periods in the labeled dataset.}}
   \label{fig:traintest}
\end{figure}

\subsection{Labeled Set}

\subsubsection{Data preparation} 

\jl{To mitigate noise caused by misalignment among DSM, LU, satellite imagery, and building footprints, particularly for small structures, we exclude buildings with an area smaller than 12.5 $m^2$. Such instances either provide insufficient visual context due to partial observations after cropping or are more sensitive to spatial misalignment and potential footprint annotation errors.} In addition, since the DSM data were collected in 2021, building footprints added after 2020 are also removed to ensure temporal consistency. The DSM and LU data are then upsampled to align with the resolution of the satellite imagery, enabling better spatial consistency.

Given that buildings with available age information are sparsely distributed across Hong Kong, we select the areas of 19 satellite images from regions with dense building distributions to construct the labeled dataset. \jl{Each image is further divided into non-overlapping patches of size 512 $\times$ 512 using a sliding window. Since not all buildings within each patch have corresponding age annotations, the labels include only buildings with known age information, while unlabeled buildings are treated as background.}

\subsubsection{Data Partitioning} 

\jl{According to the available records of the building age distribution, the building age categories are defined as Pre-1970, 1970-1980, 1980-1990, 1990-2000, 2000-2010, and 2010-2020. This temporal granularity captures stage-wise changes in the building stock while providing sufficient and relatively balanced samples within each age group for model training.} The dataset is divided into training and testing sets with a ratio of 8:2, yielding 3,358 training images and 840 testing images. The distributions of building pixels across the six time periods and the numbers of buildings in the training and testing sets are illustrated in Fig.~\ref{fig:6classes} and Fig.~\ref{fig:traintest}, respectively, which demonstrates that the dataset maintains a relatively balanced distribution across different time periods.


\begin{figure}
  \centering
   \includegraphics[width=\linewidth]{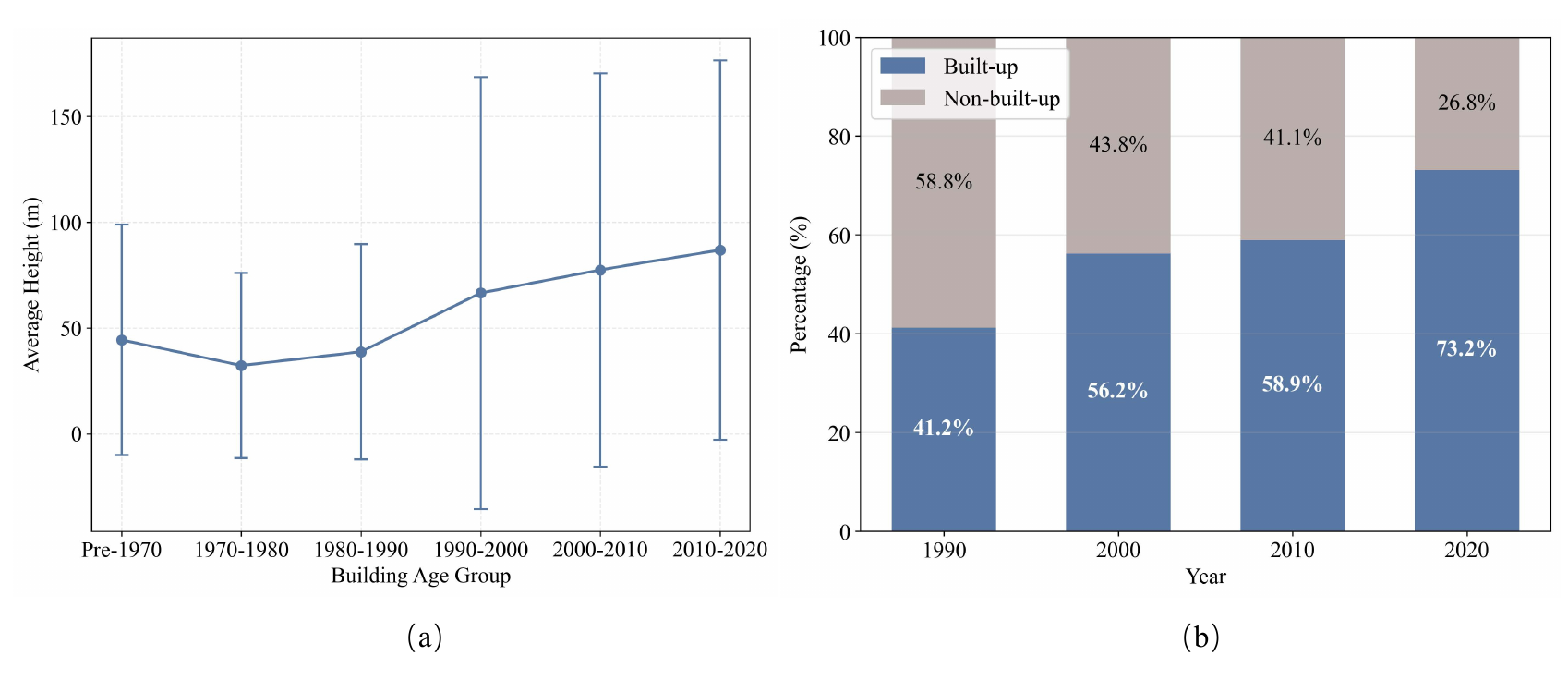}
   \caption{\jl{(a) Average building height across different construction periods. (b) The spatial-temporal changes of built-up land from 1990 to 2020.}}
   \label{fig:height_build}
\end{figure}


\subsection{Data Analysis}

\subsubsection{The relationship between building height and age}
We conduct a building height analysis across different time periods based on the available building height data. As shown in Fig.~\ref{fig:height_build} (a), buildings constructed before 1990 are relatively low-rise, with average heights ranging from approximately 32 to 44 m, whereas the mean height increases to around 87 m for buildings built between 2010 and 2020, revealing a clear upward trend over time. Meanwhile, from 1990 to 2020, the standard deviation remains relatively large in each period, indicating that many low-rise buildings still coexist with newly developed high-rise structures. \jl{From the perspective of height patterns, high-rise buildings are more likely to correspond to recent construction periods. These findings suggest that building height provides valuable and reliable cues for construction age estimation.}

\subsubsection{The relationship between the spatial-temporal changes of built-up land and building age}
We illustrate the spatial-temporal evolution of built-up land from 1990 to 2020 in Fig.~\ref{fig:height_build} (b). As shown, the share of built-up land increased dramatically from 41.2\% in 1990 to 73.2\% in 2020, while the proportion of non-built-up land declined correspondingly from 58.8\% to 26.8\%, reflecting continuous and substantial urban expansion over the past three decades. \jl{This clear temporal trend of built-up area growth provides valuable spatial-temporal prior knowledge for building age estimation, as areas that were already built in earlier years are more likely to contain older buildings, while regions converted to built-up land in later periods tend to correspond to newer constructions. These land-use change data can act as a spatial reference to constrain and refine the model’s inferences of building age, thus improving the accuracy of age estimation for individual buildings.}

\begin{table*}[h]
	\centering
	\caption{Comparison on the Hong Kong dataset of Accuracy, mF1, and mIoU at the pixel level and Accuracy, mF1, RMSE, and MAE at the instance level.}
	\setlength{\tabcolsep}{1.3mm}{
		\begin{tabular}{cc|ccc|cccc}
			\hline

                \centering \multirow{2}{1cm}{\textbf{Type}}& \multirow{2}{1cm}{\textbf{Method}}&
			         \multicolumn{3}{c|}{\textbf{Pixel}}&\multicolumn{4}{c}{\textbf{Instance}}\\
                     &&\textbf{Acc (\%) $\uparrow$}&\centering \textbf{mF1 (\%) $\uparrow$}&\textbf{mIoU (\%) $\uparrow$}&\textbf{Acc (\%) $\uparrow$}&\textbf{mF1 (\%) $\uparrow$}&\textbf{ RMSE $\downarrow$ }&\textbf{ MAE $\downarrow$}\\
                \hline
                &Unet \cite{Unet} &37.48  &35.63  &21.83  &35.80  & 32.50 & 15.5209 & 11.5319\\
                &Unetformer \cite{UNetFormer} & 51.11 & 50.07 & 33.45  & 43.85 & 42.33 & 14.9417 & 10.4037\\
                &ABCNet \cite{ABCNet} & 49.31 & 47.82 & 31.54  & 40.96 & 39.65 & 15.0633 & 10.7217 \\
                &Segformer \cite{SegFormer} & 54.73 & 53.59 & 36.77  & 48.27 & 47.02 & 13.9652 & 9.6142\\
                Image only&CMNeXt \cite{CMNeXt} & 35.68 & 35.73 & 21.99 & 33.92 & 33.97 & 15.1417 & 11.4695\\
                &TransUnet \cite{Transunet} & 31.83 & 30.82 & 18.58  & 27.70 & 26.94 & 16.1252 & 12.8558\\
                &CMTFNet \cite{CMTFNet} & 58.54 & 57.05 & 40.06  & 48.81 & 48.15 & 13.5038 & 9.3329\\
                &CMT \cite{Cmt} & 29.80 & 25.05 & 14.99 & 26.83 & 21.36 & 16.7366 & 13.2771\\
                &Deeplab \cite{deeplab} & 38.72  & 36.14 & 22.26 & 32.96 & 30.98 & 17.5092 & 13.0838 \\
                &A2FPN \cite{A2-FPN}  & 55.85 & 54.13 & 37.24 & 47.74 & 46.69 & 13.4190 & 9.3302 \\
                \hline
                 &CMX \cite{CMX} & 53.49 & 52.73 & 35.99 & 49.84 & 48.39 & 13.0639 & 8.9852\\
                Multi-modal&Asymformer \cite{Asymformer} & 41.50 & 41.35 & 32.45  & 36.61 & 36.13 & 16.4804 & 12.0615\\
                &MFNet \cite{MFNet} &46.60  & 46.40 & 30.37 & 43.13 & 41.89 & 14.1367 & 10.0224\\
                \hline
                 & Ours & \textbf{65.52} & \textbf{63.63} & \textbf{46.80} & \textbf{58.37} & \textbf{57.07} & \textbf{9.6919} & \textbf{6.6978} \\
		\hline
	\end{tabular}}
	\label{tab:Hong Kong}
\end{table*}

\section{Experiments}\label{sec:experiments}

\subsection{Datasets and Metrics}
The proposed model is evaluated on two datasets: Hong Kong and Amsterdam. The Hong Kong dataset serves as the primary dataset and contains comprehensive \jl{multi-source geospatial data} as described in detail in the previous section. The Amsterdam dataset \cite{nlba} is an open-source benchmark that includes only satellite imagery, consisting of 700 images for training, 100 for validation, and 200 for testing. According to the dataset settings, building ages are categorized into three classes: Pre-1980, 1980-2000, and Post-2000.

We evaluate the model using Accuracy, mean F1 score (mF1), mean Intersection over Union (mIoU), Mean Absolute Error (MAE), Root Mean Square Error (RMSE), where Accuracy, mF1, and mIoU are reported for pixel-level prediction, while Accuracy, mF1, MAE, and RMSE are used for instance-level evaluation. Since the Amsterdam dataset only provides building mask annotations without instance-level labels, the evaluation is performed exclusively at the pixel level. For the Hong Kong dataset, we perform a more comprehensive evaluation by reporting results at both the pixel level and the building instance level. The definition of Accuracy is given by:
\begin{equation*}
	Acc = \frac{TP}{TP+FP+TN+FN},
\end{equation*}
where TP, FP, TN, and FN represent the number of true positives, false positives, true negatives, and false negatives, respectively. mF1 and mIoU are derived from the mean F1 and mean IoU across each category. F1 and IoU are computed for each category according to the following formulas:
\begin{eqnarray}
	F1 &=& 2 \times \frac{Q \times R}{Q + R},\\
	IoU& =& \frac{TP}{TP+FN+FP},
\end{eqnarray}
where $Q$ and $R$ are given by:
\begin{eqnarray}
	Q &=&  \frac{TP}{TP+FP},\\
	R &=&  \frac{TP}{TP+FN}.
\end{eqnarray}

\jl{MAE and RMSE are computed based on the differences between the ground-truth building age $c$ and the estimated age $c'$. For the closed-ended categories, we assign the midpoint year of each category as the representative building age for evaluation. For the open-ended category, Pre-1970, we set 1965 as the representative building age. The definitions of MAE and RMSE are shown as follows:}
\begin{eqnarray}
	MAE= \frac{1}{N} \sum^{N}_{i=1} |c_i-c'_i|,
\end{eqnarray}
\begin{eqnarray}
    RMSE= \sqrt{\frac{1}{N} \sum^{N}_{i=1} (c_i-c'_i)^2}.
\end{eqnarray}

\subsection{Implementation Details}

We conduct extensive experiments using PyTorch on a single NVIDIA GeForce RTX 4090 GPU with 24 GB memory to validate the effectiveness of the proposed model. \jl{In addition to satellite imagery and building footprints, DSM and LU data are incorporated as auxiliary raster inputs, while area, length, LH ratio, and compactness, extracted from building footprints, are used as property inputs.} For the proposed CMoE module, the hyperparameter $\varepsilon$ is empirically set to 0.1, and the number of experts is set to 2. The feature dimensions within the encoder and decoder are fixed at 256 and 64, respectively. In the decoder design, the window size and dropout rate are set to 8 and 0.1, respectively, while the pyramid decoder processes multi-scale features with input resolutions of $\{16, 64, 128, 256\}$.

\begin{table}[t]
	\centering
	\caption{Comparison on the Amsterdam dataset of Accuracy, mF1, and mIoU at the pixel level.}
	\setlength{\tabcolsep}{1.3mm}{
		\begin{tabular}{c|ccc}
			\hline
                \centering \multirow{2}{1cm}{\textbf{Method}}&\multicolumn{3}{c}{\textbf{Pixel}}\\
                
                &\textbf{Acc (\%)  $\uparrow$}&\textbf{mF1 (\%)  $\uparrow$}&\textbf{mIoU (\%)  $\uparrow$}\\
                \hline
                Unet \cite{Unet} & 52.68 & 50.42 & 34.44  \\
                Unetformer \cite{UNetFormer} & 58.95 & 58.08 & 41.13  \\
                ABCNet \cite{ABCNet} & 61.57 & 60.80 & 43.99 \\
                Segformer \cite{SegFormer} & 44.32 & 34.42 & 23.36 \\
                CMNeXt \cite{CMNeXt} & 46.61 & 33.72 & 23.00 \\
                TransUnet \cite{Transunet} & 54.30 & 53.52 & 36.77 \\
                CMTFNet \cite{CMTFNet} & 62.66 & 62.33  & 45.44\\
                CMT \cite{Cmt} & 53.58 & 51.05 & 34.99 \\
                Deeplab \cite{deeplab} & 57.61  & 55.68 & 39.15 \\
                A2FPN \cite{A2-FPN}  &  60.47 & 60.02 & 43.11\\
                \hline
                 Ours & \textbf{62.68}  & \textbf{62.55} & \textbf{45.67}\\
		\hline
	\end{tabular}}
	\label{tab:Amsterdam}
\end{table}

\begin{figure}
  \centering
   \includegraphics[width=\linewidth]{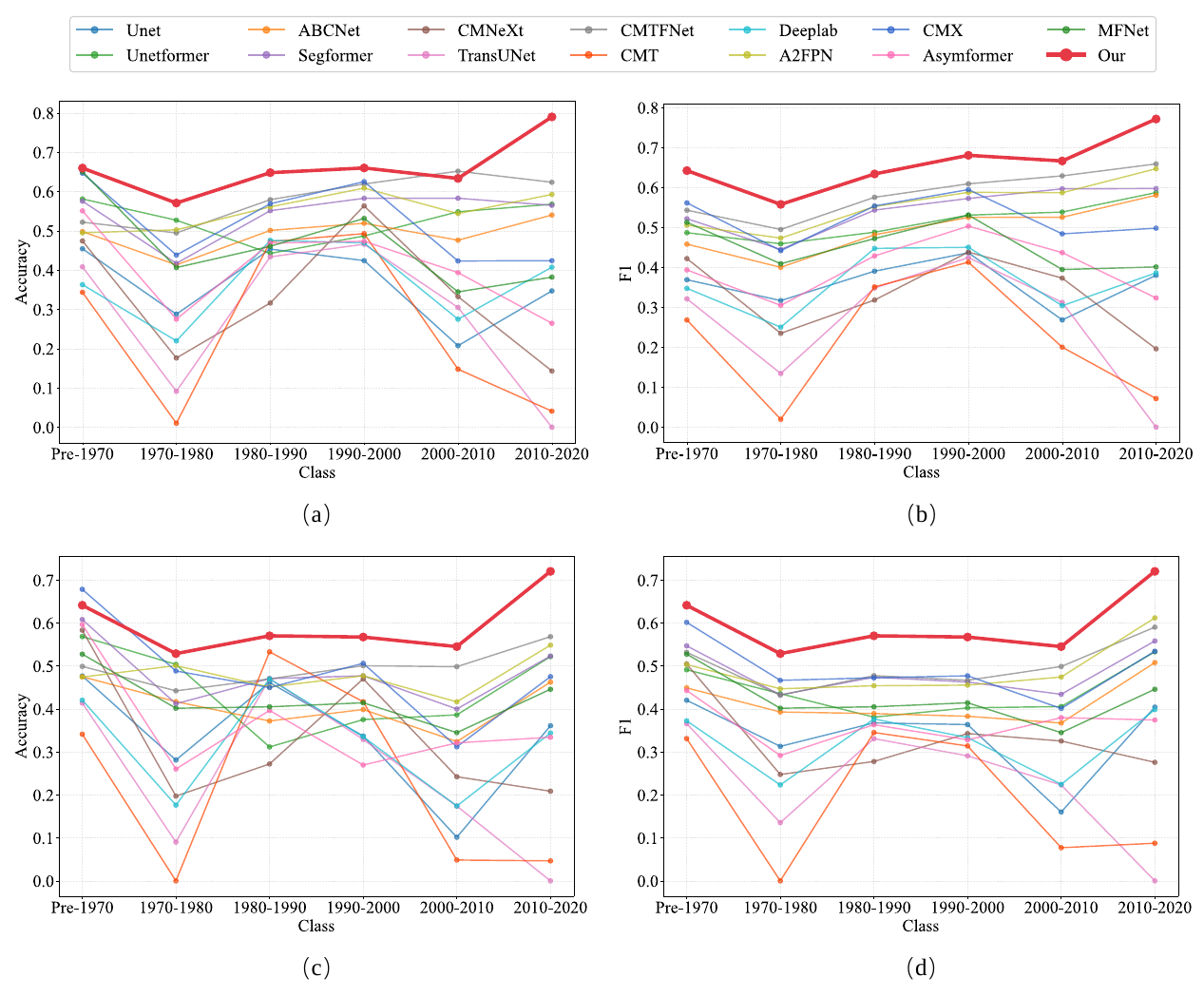}
   \caption{Performance comparison across different building age groups on the Hong Kong dataset. (a) Accuracy at the pixel level. (b) F1 at the pixel level. (c) Accuracy at the instance level. (d) F1 at the instance level.}
   \label{fig:f1acch}
\end{figure}

\subsection{Performance Comparison}

We conduct a comprehensive comparison between our model and several state-of-the-art segmentation networks, including ten image-based models, namely Unet \cite{Unet}, Unetformer \cite{UNetFormer}, ABCNet \cite{ABCNet}, Segformer \cite{SegFormer}, CMNeXt \cite{CMNeXt}, TransUnet \cite{Transunet}, CMTFNet \cite{CMTFNet}, CMT \cite{Cmt}, Deeplab \cite{deeplab}, A2FPN \cite{A2-FPN}, and three multi-modal-based models, namely CMX \cite{CMX}, Asymformer \cite{Asymformer}, MFNet \cite{MFNet}. Since the Amsterdam dataset only provides satellite images, we only apply image-based models on it. To ensure a fair evaluation, all of these methods are thoroughly optimized using our training and test datasets to achieve their peak performance.

\begin{figure}
  \centering
   \includegraphics[width=\linewidth]{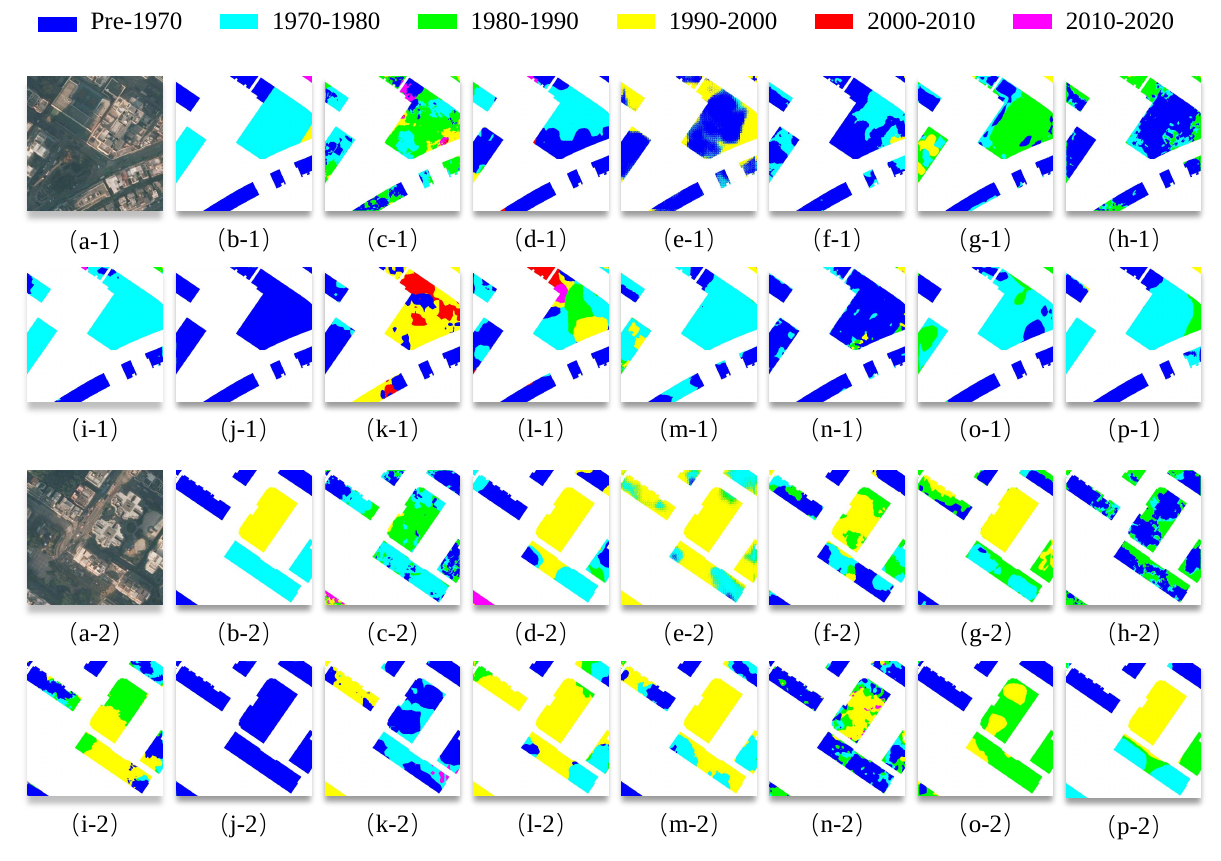}
   \caption{The visual comparisons on the Hong Kong dataset. (a) Satellite image. (b) Ground truth. (c) Unet. (d) Unetformer.  (e) ABCNet. (f) Segformer. (g) CMNeXt. (h) TransUnet. (i) CMTFNet. (j) CMT. (k) Deeplab. (l) A2FPN. (m) CMX. (n) Asymformer. (o) MFNet. (p) Ours.}
   \label{fig:hongkongresult}
\end{figure}

\begin{figure}
  \centering
   \includegraphics[width=\linewidth]{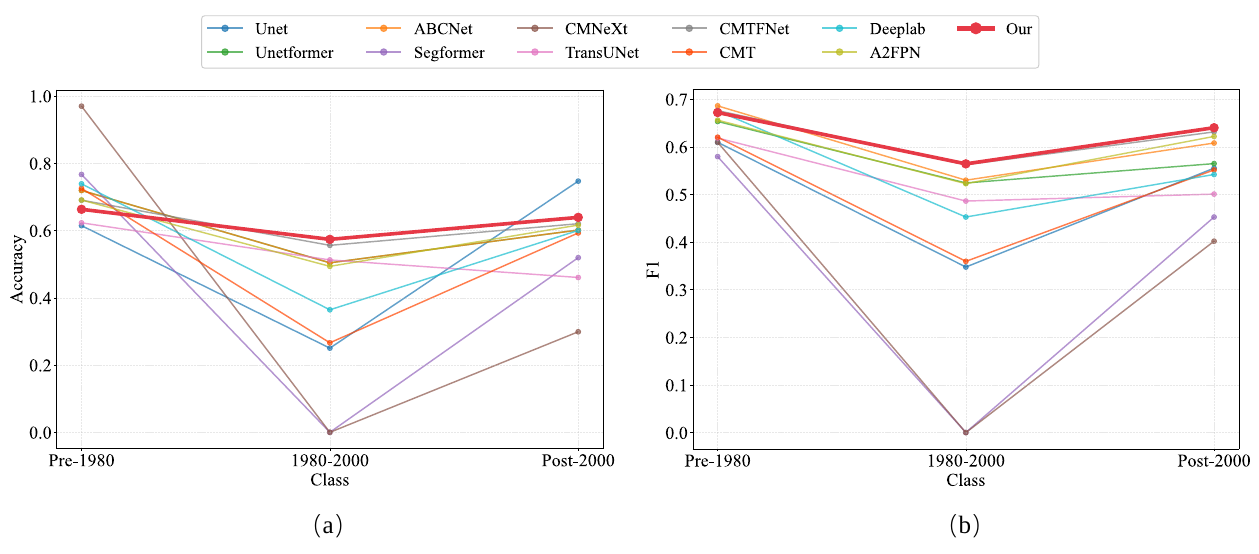}
   \caption{Performance comparison across different building age groups on the Amsterdam dataset. (a) Accuracy at the pixel level. (b) F1 at the pixel level.}
   \label{fig:f1acca}
\end{figure}

\begin{figure}
  \centering
   \includegraphics[width=0.9\linewidth]{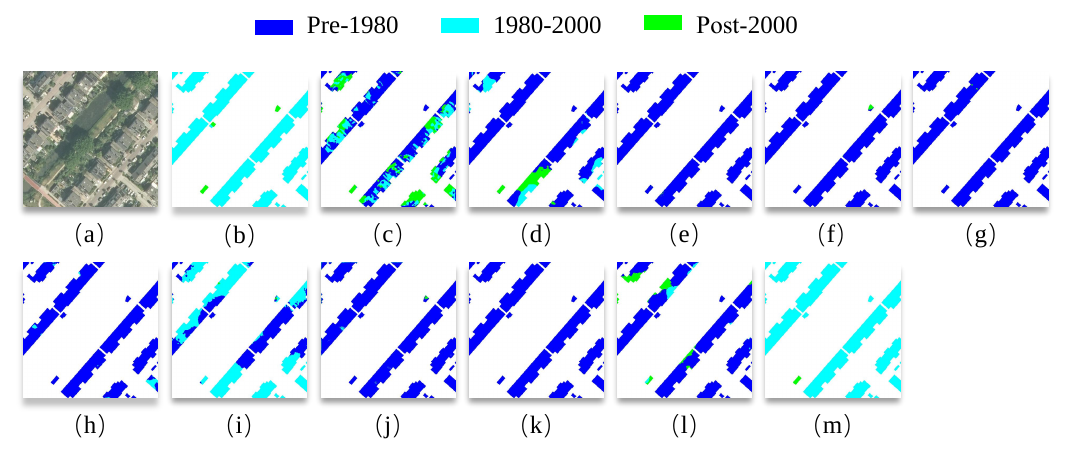}
   \caption{\jl{The visual comparisons on the Amsterdam dataset.  (a) Satellite image. (b) Ground truth. (c) Unet. (d) Unetformer.  (e) ABCNet. (f) Segformer. (g) CMNeXt. (h) TransUnet. (i) CMTFNet. (j) CMT. (k) Deeplab. (l) A2FPN. (m) Ours.}}
   \label{fig:amsterdamresult}
\end{figure}

The quantitative results of these experiments on the Hong Kong and Amsterdam datasets are presented in Table~\ref{tab:Hong Kong} and Table~\ref{tab:Amsterdam}, respectively. The detailed metrics for each class and the visual comparisons for Hong Kong and Amsterdam are shown in Figs.~\ref{fig:f1acch}-\ref{fig:amsterdamresult}.

\subsubsection{Hong Kong}

Table~\ref{tab:Hong Kong} presents the results of all models on the Hong Kong dataset, where our model demonstrates significantly better performance than conventional segmentation methods. This improvement can be attributed to the targeted design of our \jl{multi-source} information fusion framework, particularly in instance-level estimation. Compared with the second-best method, CMTFNet, our model achieves improvements of 7\%, 6\%, and 6\% in Accuracy, mF1, and mIoU for pixel-level estimation, respectively. For the instance level, it also shows gains of 10\%, 9\%, and reductions of 4 and 3 in RMSE and MAE, respectively. Furthermore, the lower RMSE and MAE values indicate that our model better captures the ordinal nature of the building age estimation task, effectively minimizing the discrepancy between predicted age and ground truth. In addition, multi-modal approaches such as CMX, AsymFormer, and MFNet consistently outperform most image-based methods, further demonstrating the effectiveness of integrating \jl{multi-source} information for this task.

As shown in Fig.~\ref{fig:f1acch}, our model achieves consistently strong performance across all classes in both Accuracy and F1 score. The performance gap between different classes is also relatively small, indicating that the model effectively learns the intrinsic characteristics of each class rather than overfitting to specific categories. In contrast, models such as CMT, TransUNet, and CMNeXt exhibit more uneven performance across classes, particularly showing inferior results on the 1970-1980 and 2010-2020 categories.

Fig.~\ref{fig:hongkongresult} presents the visual comparison of pixel-level estimation results between our model and competing methods. Due to the lack of instance-level constraints in pixel-wise prediction, many methods produce noisy predictions within building instances. In comparison, our model generates smoother and more consistent results with significantly reduced noise, and its predictions are more closely aligned with the ground truth. Models such as CMT and ABCNet tend to assign most pixels within a building to only one or two dominant age categories. This behavior highlights their limitation in capturing fine-grained age clustering patterns across different building instances.

\subsubsection{Amsterdam}

As shown in Table~\ref{tab:Amsterdam}, our model still achieves the best results among all the metrics in this dataset, demonstrating its effectiveness not only in the difficult task but also in the relatively simple task. Since there are only three classes in the Amsterdam dataset, the difference between the performance of our model and the other models is not large compared with the Hong Kong dataset. Most models perform well, such as ABCNet, CMTFNet, and A2FPN, which achieve more than 60\% in Accuracy and mF1. \jl{Compared with the Amsterdam and Hong Kong datasets, most compared models achieve better performance on Amsterdam but degrade significantly on Hong Kong, while our model maintains robust performance. This demonstrates the effectiveness of incorporating additional raster and property data in handling challenging urban scenes.}

\jl{As shown in Fig.~\ref{fig:f1acca}, most models perform well on the Pre-1980 and Post-2000 classes due to their larger sample sizes and clearer patterns, but struggle with the 1980-2000 class. In contrast, our model achieves balanced performance across all categories without a significant drop.}

\jl{Fig.~\ref{fig:amsterdamresult} shows the visual comparison on the Amsterdam dataset. Most competing methods tend to misclassify 1980-2000 buildings as Pre-1980, reflecting a bias toward dominant classes. Our method better distinguishes this category and accurately predicts both 1980-2000 buildings and small Post-2000 buildings, demonstrating its robustness.}

\subsection{Feasibility of filling the missing data}
\begin{figure}
  \centering
   \includegraphics[width=\linewidth]{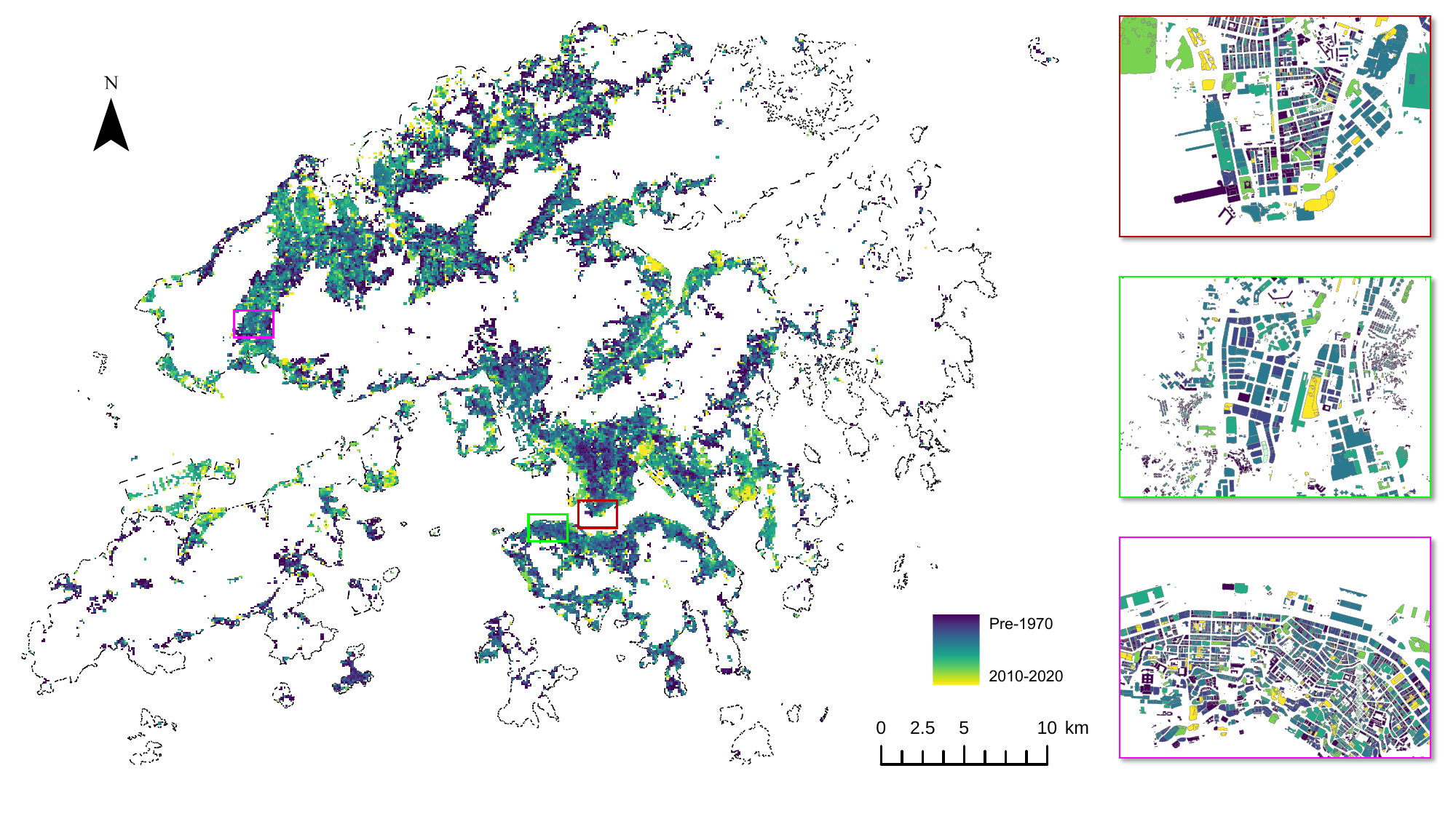}
   \caption{\jl{The left figure shows the overall distribution of building ages across Hong Kong by aggregating footprint-level age information into 100 m spatial grids, while the three subfigures on the right illustrate building age distributions in three representative areas at the individual building instance level. To visually represent these aggregated values, the six predefined age categories are evenly distributed along a continuous color bar, where the gradual color transitions between adjacent categories reflect the continuous intermediate category values generated through spatial aggregation rather than an exact continuous age. In contrast, the three subfigures on the right show individual building footprints, where each building is assigned to one of the six discrete age categories and no intermediate values are generated.}}
   \label{fig:hongkong_age}
\end{figure}

\begin{figure}
  \centering
   \includegraphics[width=\linewidth]{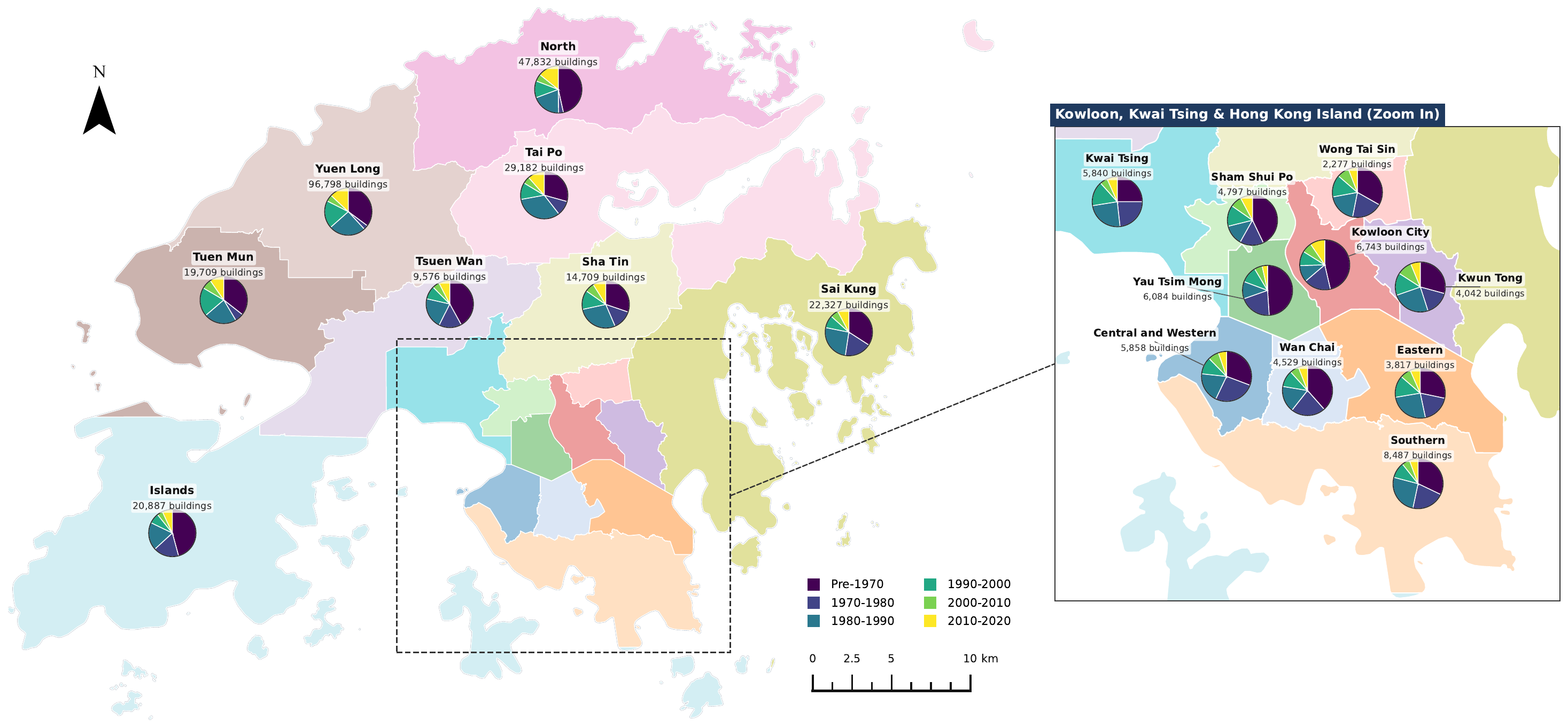}
   \caption{Building age composition and building stock across the 18 districts of Hong Kong. Each pie chart shows the share of a district's buildings falling into each construction-period class (Pre-1970 to 2010-2020), and the label beneath the district name gives its total number of buildings. The densely built urban core (Kowloon and Hong Kong Island) is enlarged in the panel on the right.}
   \label{fig:hongkong_district}
\end{figure}

\begin{figure}
  \centering
   \includegraphics[width=\linewidth]{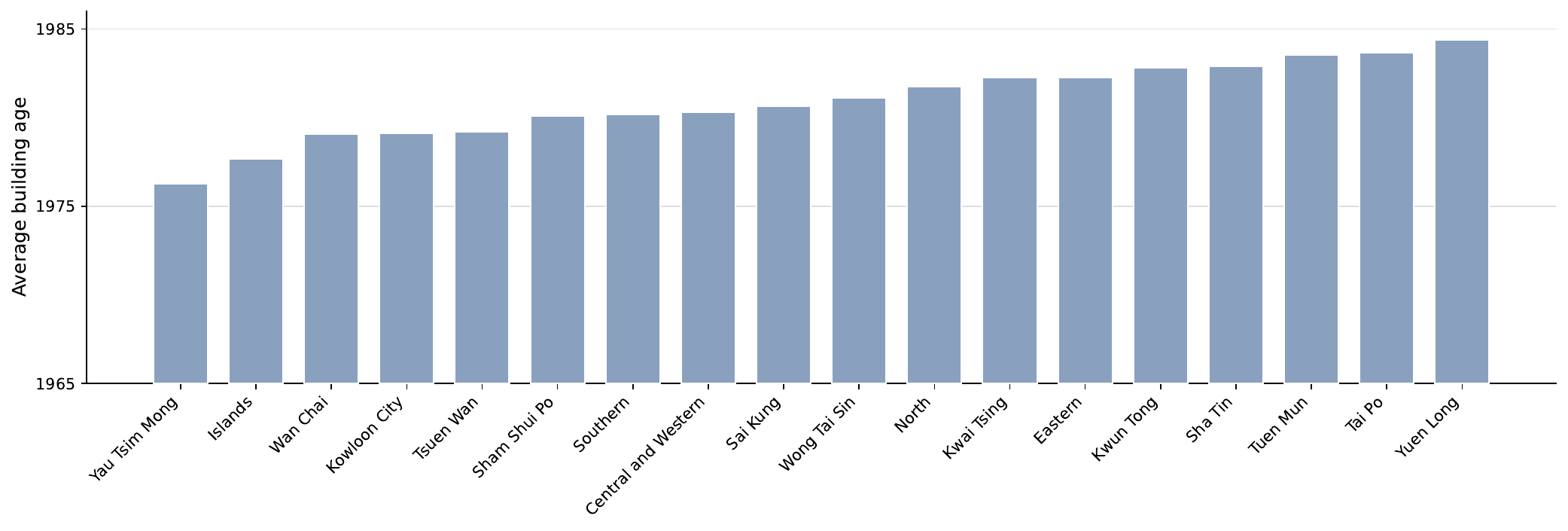}
   \caption{Average building age by district. Bars show the average construction year of each district's buildings. A lower value therefore indicates an older building stock. Districts are ordered from the oldest (left) to the most recent (right).}
   \label{fig:mean_class}
\end{figure}

To demonstrate the effectiveness of the proposed model for city-scale analysis, we apply the model pretrained on the labeled dataset to the entire Hong Kong region. For model inference over the entire Hong Kong region, we adopt an overlapping cropping strategy to generate image patches. Specifically, the remote sensing images are cropped into 512 $\times$ 512 patches with a stride of 256. This strategy allows buildings to appear in multiple patches and preserves sufficient spatial context for instance-level prediction. For buildings appearing in multiple patches, the predictions from different patches are aggregated using the corresponding building areas within each patch as weighting factors, and the weighted aggregation results are used as the final instance-level predictions. In this process, DSM and LU data are incorporated as additional raster inputs, while building attributes, including area, length, LH ratio, and compactness, are used as property features. The model configuration and input data settings are consistent with those used during training. The resulting spatial distribution of building ages in Hong Kong is illustrated in Fig.~\ref{fig:hongkong_age}. The left figure shows the overall distribution of building ages across Hong Kong by aggregating footprint-level age information into 100 m spatial grids, while the three subfigures on the right illustrate building age distributions in three representative areas at the individual building instance level. As shown in the three subfigures, our method actually assigns a discrete age category to each individual building footprint across the entire Hong Kong area. However, raw footprint units are overly fragmented, making it difficult to visualize the overall city-scale distribution directly. To resolve this visualization issue, we aggregate the footprint-level age information into 100 m spatial grids. The value of each grid is calculated by averaging the predicted age categories of all building pixels within that grid. As a result, the averaging process generates continuous intermediate values between the predefined age categories, representing the overall age characteristics of buildings within each grid. Furthermore, Fig.\ref{fig:hongkong_district} presents the building age composition and building stock across Hong Kong's 18 districts, with each district's average building age summarised in Fig.~\ref{fig:mean_class}, which enables a better analysis of the differences across Hong Kong's districts.

As shown in Fig.~\ref{fig:hongkong_age}, the fine-grained spatial distribution of building ages reveals substantial heterogeneity across Hong Kong. Kowloon exhibits a particularly dense urban fabric, with buildings occupying almost the entire region, whereas the New Territories show a more dispersed distribution shaped by local topography and geographical constraints. The spatial distribution of building ages also reflects different stages of urban development. In the northern part of the Outlying Islands, relatively new buildings are prominent, consistent with recent land reclamation and urban expansion. In contrast, older buildings are more prevalent in established urban areas, while coastal areas generally exhibit a greater concentration of newer buildings than inland areas. These patterns highlight the spatially uneven development of Hong Kong's building stock and provide a fine-grained view of its urban evolution. To further examine these spatial variations at the administrative level, Fig.\ref{fig:hongkong_district} presents the building age composition across Hong Kong's 18 districts, while Fig.\ref{fig:mean_class} compares their average building age. Yau Tsim Mong has the largest proportion of Pre-1970 buildings, with nearly half of its building stock belonging to this category. Yuen Long, by contrast, exhibits a more diverse and bimodal age composition, combining substantial concentrations of both Pre-1970 buildings and buildings constructed during 1980-1990 and 2010-2020. The district-level average building age further highlights these contrasts, as Yau Tsim Mong, the Islands and Wan Chai show the oldest average profiles while Yuen Long, Tai Po and Tuen Mun are the most recent, reflecting the substantial contribution of later-period construction. Taken together, the district-level statistics complement the fine-grained spatial patterns by quantifying differences in building age composition across administrative units. This multiscale analysis demonstrates the potential of the proposed framework to support urban development assessment at both local and district levels.


\subsection{Parameter Analysis}

\begin{table*}[h]
	\centering
	\caption{\jl{Parameter Analysis of encoder and decoder dimensions on the Hong Kong dataset}}
	\setlength{\tabcolsep}{1.3mm}{
		\begin{tabular}{cc|ccc|cccc}
			\hline
                \centering \multirow{2}{1cm}{ \centering\textbf{Type}}& \multirow{2}{1cm}{ \centering \textbf{Dim}}&
			         \multicolumn{3}{c|}{\textbf{Pixel}}&\multicolumn{4}{c}{\textbf{Instance}}\\
                     &&\textbf{Acc (\%) $\uparrow$}&\centering \textbf{mF1 (\%) $\uparrow$}&\textbf{mIoU (\%) $\uparrow$}&\textbf{Acc (\%) $\uparrow$}&\textbf{mF1 (\%) $\uparrow$}&\textbf{ RMSE $\downarrow$ }&\textbf{ MAE $\downarrow$}\\
                \hline
                 \multirow{4}{1cm}{Encoder}& 64 & \textbf{65.63} & \textbf{64.07} & \textbf{47.36} & 57.52 & 56.46 &  9.8087&  6.7797 \\
                & 128 & 65.52 & 63.87 & 47.09 & 57.75 & 56.73 & 9.8979 & 6.8771 \\
               &256 & 65.52 & 63.63 & 46.80 & \textbf{58.37} & \textbf{57.07} & \textbf{9.6919} & \textbf{6.6978}\\
                & 512 & 64.37 & 62.29 & 45.42 & 56.14 & 54.95 & 9.9063 & 6.9569 \\
                \hline
                 \multirow{4}{1cm}{Decoder}&16  & 63.87 & 61.86 & 44.99 & 55.57 & 53.96 & 10.0860 & 7.0862\\
                &32 & 64.97 & 63.37 & 46.60 & 57.14 & 56.35 & 10.0861 &  6.9626\\
                &64 & \textbf{65.52} & \textbf{63.63} & \textbf{46.80} & \textbf{58.37} & \textbf{57.07} & \textbf{9.6919} & \textbf{6.6978}\\
                &128   & 64.03 & 62.71 & 45.85 & 56.95 & 55.88 & 9.8425 & 6.8819 \\
		\hline
	\end{tabular}}
	\label{tab:encoderdecoder}
\end{table*}
\subsubsection{Encoder dimension}

The first part of Table~\ref{tab:encoderdecoder} presents the parameter analysis of encoder feature dimensions of $\{64,128,256,512\}$. For pixel-level prediction, the best performance is achieved at 64 dimensions, while larger dimensions provide no further benefit, suggesting that compact features are sufficient to capture local semantic patterns. In contrast, instance-level performance improves as the dimension increases from 64 to 256 dimensions, reaching its peak at 256 dimensions. This indicates that building-level age estimation requires richer representations to model global structural and contextual information. When the dimension is further increased to 512 dimensions, performance declines slightly, likely due to redundancy and optimization difficulty. These results highlight the different encoder feature requirements of pixel-level and instance-level tasks.

\subsubsection{Decoder dimension}

In the second part of Table~\ref{tab:encoderdecoder}, we evaluate the model with decoder dimensions of $\{16,32,64,128\}$. As the decoder dimension increases from 16 to 64, both pixel-level and instance-level metrics improve consistently, indicating that richer decoder feature representations are beneficial for prediction. The best performance is achieved at 64 dimensions, where the model attains the most balanced results across all evaluation metrics. However, when the dimension is further increased to 128, performance declines slightly, suggesting that excessive parameterization may reduce the effectiveness of feature decoding.

Overall, these results demonstrate that moderate feature dimensions are more suitable for the proposed framework, while excessively small dimensions limit representation capacity and overly large dimensions may introduce redundancy. In general, although the final performance varies under different parameter settings, the overall differences remain relatively small and stable, demonstrating the robustness of the proposed framework to changes in feature dimensions.

\subsection{CMoE Activation Visualization}

\begin{figure}
  \centering
   \includegraphics[width=0.5\linewidth]{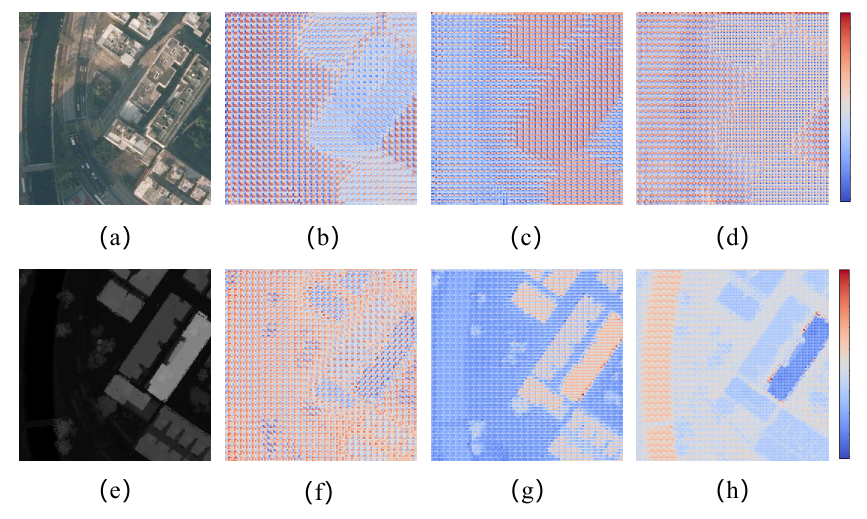}
   \caption{The visual comparisons of cosine similarity features and expert responses in the CMoE module with and without DSM data. (a) Satellite image. (b) Features after the cosine similarity without DSM. (c) Features after the first expert without DSM. (d) Features after the second expert without DSM. (e) DSM data. (f) Features after the cosine similarity with DSM. (g) Features after the first expert with DSM. (h) Features after the second expert with DSM.}
   \label{fig:moev}
\end{figure}

Fig.~\ref{fig:moev} presents a visual comparison of the cosine similarity features and expert responses in the CMoE module, with and without DSM data. Specifically, Fig.~\ref{fig:moev} (b), (c), and (d) show the outputs without DSM, while Fig.~\ref{fig:moev} (f), (g), and (h) illustrate the corresponding results with DSM. By comparing Fig.~\ref{fig:moev} (b) and (f), it can be observed that the cosine similarity mechanism adaptively adjusts feature weights depending on whether additional DSM data are provided, producing distinct feature responses under varying data conditions. Furthermore, comparisons between Fig.~\ref{fig:moev} (c) and (d), as well as Fig.~\ref{fig:moev} (g) and (h), reveal that different experts learn complementary feature distributions. For instance, the first expert in Fig.~\ref{fig:moev} (c) and (g) primarily emphasizes building regions while suppressing background responses, whereas the second expert in Fig.~\ref{fig:moev} (d) and (h) exhibits the opposite tendency. These visualizations demonstrate that the cosine similarity mechanism effectively adapts \jl{multi-source data} feature weighting, while the expert branches decompose the fusion task into specialized subproblems, thereby enhancing representation learning.

\subsection{Effect of Pixel Count on Age Estimation}
\begin{figure}
  \centering
   \includegraphics[width=0.6\linewidth]{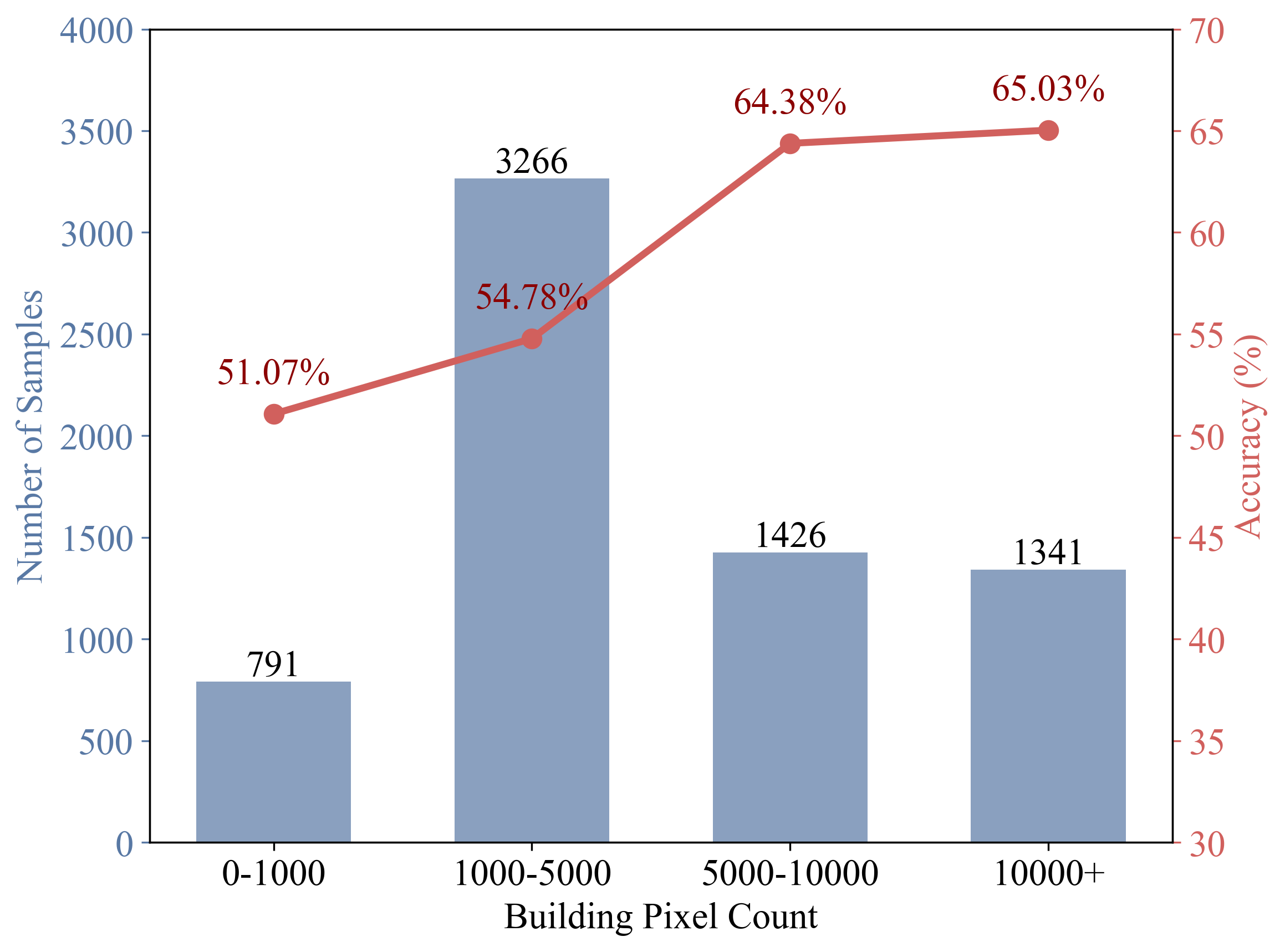}
   \caption{Accuracy distribution across different building pixel counts.}
   \label{fig:accuracy_count_distribution}
\end{figure}

Fig.~\ref{fig:accuracy_count_distribution} illustrates the relationship between building pixel count and instance-level accuracy in the test set. Buildings are categorized into four groups $\{0\text{-}1000,1000\text{-}5000,5000\text{-}10000,10000+\}$ according to their pixel coverage. As shown in Fig.~\ref{fig:accuracy_count_distribution}, the majority of building instances fall within the 1000-5000 pixel range, while the numbers of buildings in the 5000-10000 and 10000+ pixel groups are approximately comparable. A clear increasing trend in estimation accuracy is observed as the number of building pixels grows. Specifically, the model achieves accuracies of 51.07\% and 54.78\% for buildings with 0-1000 and 1000-5000 pixels, respectively, while the performance improves to 64\%-65\% when the pixel count exceeds 5000. These results indicate a strong positive relationship between building size and prediction accuracy. \jl{Buildings that become small after cropping contain fewer spatial structures, textures, and contextual cues, resulting in weaker feature representations and lower prediction performance. For inherently small buildings, they not only provide fewer visual cues for age prediction but are also more sensitive to spatial misalignment between raster images and building footprints, which can further affect prediction accuracy.}

\begin{table*}[h]
	\centering
	\caption{Impact of DSM at different proportions on age estimation performance.}
	\setlength{\tabcolsep}{1.3mm}{
		\begin{tabular}{c|ccc|cccc}
			\hline
                \multirow{2}{1cm}{\textbf{Ratio}}&
			         \multicolumn{3}{c|}{\textbf{Pixel}}&\multicolumn{4}{c}{\textbf{Instance}}\\
                     &\textbf{Acc (\%) $\uparrow$}&\centering \textbf{mF1 (\%) $\uparrow$}&\textbf{mIoU (\%) $\uparrow$}&\textbf{Acc (\%) $\uparrow$}&\textbf{mF1 (\%) $\uparrow$}&\textbf{RMSE $\downarrow$}&\textbf{MAE $\downarrow$}\\
                \hline
                0\%& 63.45 & 61.86 & 44.94 & 54.94 & 53.77 & 10.6374 & 7.3914\\
               20\% &  63.67 & 62.08 & 45.17 & 55.28 & 54.08 & 10.5328 & 7.3158 \\
               40\% & 63.80 & 62.22 & 45.32 & 55.54 & 54.36 & 10.5312 & 7.3077 \\
            60\% & 64.21 & 62.61 & 45.73 & 56.59 & 55.35 & 10.2738 & 7.1202\\
            80\% & 64.45 & 62.78 & 45.91 & 56.93 & 55.57 & 10.2249 &  7.0769\\
           100\% & 64.66 & 62.98 & 46.12 & 57.33 & 55.94 &  10.0731 &  6.9767 \\
            Complete& \textbf{65.52} & \textbf{63.63} & \textbf{46.80} & \textbf{58.37} & \textbf{57.07} & \textbf{9.6919} & \textbf{6.6978}\\
		\hline
	\end{tabular}}
	\label{tab:DSMratio}
\end{table*}

\begin{table*}[h]
	\centering
	\caption{\jl{Ablation study of multi-source geospatial data on the Hong Kong dataset.}}
	\setlength{\tabcolsep}{1.3mm}{
		\begin{tabular}{c|ccc|cccc}
			\hline

                \multirow{2}{1cm}{\textbf{Method}}&
			         \multicolumn{3}{c|}{\textbf{Pixel}}&\multicolumn{4}{c}{\textbf{Instance}}\\
                     &\textbf{Acc (\%) $\uparrow$}&\centering \textbf{mF1 (\%) $\uparrow$}&\textbf{mIoU (\%) $\uparrow$}&\textbf{Acc (\%) $\uparrow$}&\textbf{mF1 (\%) $\uparrow$}&\textbf{RMSE $\downarrow$}&\textbf{MAE $\downarrow$}\\
                \hline
                DSM + Property  & 43.71 & 44.87 & 29.06 & 42.61 & 44.20 & 12.0516 & 8.9089 \\
               Image only & 62.47 & 59.84 & 43.06 & 53.91 & 52.86 & 10.9541 & 7.6076\\
               Image + Property & 63.30 & 61.71 & 44.77 & 54.63 & 53.50 & 10.6200 & 7.3926\\
            Image + DSM + Property  & 65.01 & 63.14 & 46.29 & 57.88 & 56.64 & 9.7749 & 6.7446\\
            Image + LU + Property& 63.45 & 61.86 & 44.94 & 54.94 & 53.77 & 10.6374 & 7.3914\\
            Image + DSM + LU + Property & \textbf{65.52} & \textbf{63.63} & \textbf{46.80} & \textbf{58.37} & \textbf{57.07} & \textbf{9.6919} & \textbf{6.6978}\\
		\hline
	\end{tabular}}
	\label{tab:ablation}
\end{table*}

\subsection{\jl{Instance-level generalization analysis}}

\begin{table*}[h]
	\centering
	\caption{\jl{Performance evaluation under different data subsets, where Number denotes the number of buildings.}}
	\setlength{\tabcolsep}{1.3mm}{
		\begin{tabular}{cc|ccc|cccc}
			\hline
                \multirow{2}{1cm}{\textbf{Set}}&\multirow{2}{1.3cm}{\textbf{Number}}&
			         \multicolumn{3}{c|}{\textbf{Pixel}}&\multicolumn{4}{c}{\textbf{Instance}}\\
                     &&\textbf{Acc (\%) $\uparrow$}&\centering \textbf{mF1 (\%) $\uparrow$}&\textbf{mIoU (\%) $\uparrow$}&\textbf{Acc (\%) $\uparrow$}&\textbf{mF1 (\%) $\uparrow$}&\textbf{RMSE $\downarrow$}&\textbf{MAE $\downarrow$}\\
                \hline
                Training set & 23,134  & 82.00 & 80.92 & 68.22 & 67.75 & 67.17 & 8.5639 & 5.8254 \\
              Filtered test set & 3,361 & 61.10 & 60.35 & 43.28 & 59.12 & 58.31 & 9.4499 & 6.5159 \\
            Full test set & 6,824 & 65.52 & 63.63 & 46.80 & 58.37 & 57.07 & 9.6919 & 6.6978\\
		\hline
	\end{tabular}}
	\label{tab:overlap}
\end{table*}

\jl{Since our remote sensing images are divided into non-overlapping patches and randomly assigned to the training and test sets, a single building may be split into multiple patches, with different parts potentially appearing in the two sets. To further verify that the effectiveness of our method does not rely on memorizing local appearance information from partially observed buildings in the training set, we conduct an additional experiment by removing from the test set all buildings which also appear in the training set. The results on the training set, the filtered test set, and the original test set are presented in Table~\ref{tab:overlap}. The higher performance on the training set compared with the test set is expected and reflects the normal generalization gap between the train set and test set. The filtered test set still contains a sufficient number of building instances for evaluation, and the model maintains comparable instance-level performance after removing potentially overlapping buildings. Although the pixel-level metrics show a decrease, the filtered test set still achieves reasonable overall performance. This indicates that the performance of our method is not substantially affected by the specific training-test association, and that the random patch-based split still provides a fair evaluation of the model performance. The strong performance of our model reflects its ability to learn age-related visual and structural patterns and generalize to unseen building instances, rather than being a result of the current training-test split.}

\subsection{\jl{Cross-region generalization analysis}}
\begin{table*}[h]
	\centering
	\caption{\jl{Comparison on the Hong Kong dataset under spatial split.}}
	\setlength{\tabcolsep}{1.3mm}{
		\begin{tabular}{c|ccc|cccc}
			\hline
                \centering  \multirow{2}{1cm}{\textbf{Method}}&
			         \multicolumn{3}{c|}{\textbf{Pixel}}&\multicolumn{4}{c}{\textbf{Instance}}\\
                     &\textbf{Acc (\%) $\uparrow$}&\centering \textbf{mF1 (\%) $\uparrow$}&\textbf{mIoU (\%) $\uparrow$}&\textbf{Acc (\%) $\uparrow$}&\textbf{mF1 (\%) $\uparrow$}&\textbf{ RMSE $\downarrow$ }&\textbf{ MAE $\downarrow$}\\
                \hline
                Unetformer \cite{UNetFormer} & 36.84 & 33.44 & 20.34 & 34.48 & 31.04 & 17.3992 & 12.8084 \\
                Segformer \cite{SegFormer} & 34.23 & 32.85 & 19.78  & 33.66 & 31.56 & 18.1527 & 13.4193 \\
                CMTFNet \cite{CMTFNet} & 36.16 & 34.48 & 21.03  & 35.48 & 33.87 & 17.0559 & 12.4664 \\
                A2FPN \cite{A2-FPN}  & 35.95 & 33.85 & 20.54  & 34.84  & 32.59 & 17.6333 & 12.8944\\
                CMX \cite{CMX} & 32.69 & 31.92 & 19.26 & 32.61 & 30.22 & 18.7380 & 13.9637 \\
                Ours & \textbf{42.96} & \textbf{40.77} & \textbf{25.80} & \textbf{41.38} & \textbf{39.59} & \textbf{14.0646} & \textbf{10.0541} \\
		\hline
	\end{tabular}}
	\label{tab:spatial}
\end{table*}

\jl{We conduct an experiment in which the training and test sets are strictly separated by spatial regions. Specifically, among the 19 remote sensing images, 4 geographically independent images are selected as the test set, while the remaining 15 images are used for training. Except for the spatial partitioning strategy, all other experimental settings remain the same as those in the original experiments. We evaluate our method and compare it with several representative methods that achieve strong performance in the original experiments. As shown in Table~\ref{tab:spatial}, the performance decreases noticeably under this spatially separated setting compared with the original random split. This is consistent with the design of our framework, which benefits from the local spatial context provided by nearby labeled buildings. Once this contextual information is removed, the prediction task becomes substantially more challenging. This observation indicates that transferring the proposed framework to regions without labeled reference buildings may result in reduced performance. Nevertheless, our method still achieves the best performance at both pixel and instance levels, demonstrating that the proposed framework remains effective even under stricter spatial generalization settings.}

\subsection{Ablation Study}

\subsubsection{\jl{Multi-source geospatial data}}
The ablation results under varying proportions of DSM auxiliary data are summarized in Table~\ref{tab:DSMratio}. The first six rows report the performance under different DSM data ratios, where a modality masking strategy is applied during training with DSM availability set to 70\%, i.e., DSM is randomly dropped with a probability of 30\%. The final row presents the performance of the model trained with fully available DSM data. Overall, model performance exhibits a consistent upward trend as the proportion of complete DSM data increases, across both pixel-level and instance-level age estimation metrics. Specifically, as the DSM injection ratio increases from 0\% to 100\%, pixel-wise Accuracy, mF1, and mIoU improve steadily by 1.2\%, 2.8\%, and 1.2\%, respectively. Moreover, instance-level metrics are progressively enhanced, while regression errors, RMSE, and MAE are consistently reduced by 0.6 and 0.4, respectively. These results indicate that DSM provides effective auxiliary prior knowledge, and increasing its completeness further enhances building age estimation accuracy by offering richer and more informative reference cues. Nevertheless, a comparison between the last two rows reveals a trade-off between performance and robustness to missing modalities. Training the model to handle incomplete DSM inputs introduces a slight degradation in model accuracy. In contrast, models trained with fully available DSM data achieve superior performance, highlighting that maximizing input completeness remains critical for optimal results.

The ablation results in Table~\ref{tab:ablation} demonstrate the effectiveness of \jl{multi-source geospatial data} for building age estimation. \jl{A comparison between the first two rows reveals a clear performance degradation in the absence of imagery, indicating that the predictions are primarily driven by visual features. Compared with the second and third rows, property information yields consistent improvements, highlighting the value of complementary semantic attributes.} \jl{More substantial gains are achieved with the inclusion of DSM, where pixel-level Accuracy increases by approximately 3.0\%, while RMSE and MAE decrease by about 1.2 and 3.0, respectively, highlighting the importance of geometric and height-related cues. The incorporation of LU information yields further improvements by providing richer contextual understanding.} The best performance is achieved when all modalities are jointly exploited, with pixel-level Accuracy reaching 65.52\%, and instance-level Accuracy improving to 58.37\%, alongside further reductions in RMSE and MAE. Overall, these results indicate that \jl{multi-source geospatial data} provide complementary geometric, semantic, and contextual information, and their effective integration leads to more accurate and reliable building age estimation.

\subsubsection{Framework architecture}
Table~\ref{tab:ablation2} presents the ablation study on the proposed framework and loss functions. In the first line, without using the instance-level estimation module, instance age is obtained by directly assigning the dominant pixel-level class within its mask. In the second line, only the instance-level loss is calculated, without generating or constraining pixel-level results. In the third line, the proposed TOCL is replaced with the standard cross-entropy loss. As shown in Table~\ref{tab:ablation2}, neither pixel-level nor instance-level learning alone achieves satisfactory results. In particular, using only instance-level supervision performs even worse than deriving instance-level predictions from pixel-level results. This is because many buildings in remote sensing images are only partially visible, making it difficult for incomplete instance features to establish reliable associations with building age without the prior knowledge and constraints provided by pixel-level supervision. The combination of pixel-level and instance-level estimation modules substantially improves all evaluation metrics, achieving the highest Accuracy and mF1 scores, with gains of 0.6\% and 0.7\% over the pixel-level-only setting, and 3.0\% and 2.9\% over the instance-level-only setting, respectively. This demonstrates that pixel-level and instance-level constraints complement each other and jointly improve the final age estimation performance. \jl{Furthermore, incorporating TOCL slightly reduces Accuracy and mF1, but significantly improves RMSE and MAE by 0.6 and 0.2, respectively. This is because TOCL emphasizes the ordinal continuity of building ages, thereby enhancing chronological consistency and reducing large prediction deviations.}

\subsubsection{Model modules}

\begin{table}[t]
	\centering
	\caption{Ablation study of the whole framework. PEM and IEM represent the pixel-level and instance-level estimation modules, respectively.}
	\setlength{\tabcolsep}{1.3mm}{
		\begin{tabular}{ccc|cccc}
			\hline

             \multirow{2}{1cm}{ \centering \textbf{PEM}}&\multirow{2}{1cm}{ \centering \textbf{IEM}}&\multirow{2}{1cm}{\centering \textbf{TOCL}}&\multicolumn{4}{c}{\textbf{Instance}}\\
                     &&&\textbf{Acc (\%) $\uparrow$}&\textbf{mF1 (\%) $\uparrow$}&\textbf{RMSE $\downarrow$}&\textbf{MAE $\downarrow$}\\
                \hline
                \checkmark & &  & 58.70 & 57.36 & 10.4868 & 7.0139\\
                 & \checkmark &  & 56.33 & 55.13 & 11.2805 & 7.6128\\
                \checkmark&\checkmark & & \textbf{59.36} & \textbf{58.02} & 10.2441 & 6.8792 \\
                 \checkmark &\checkmark & \checkmark & 58.37 & 57.07  &\textbf{9.6919}&\textbf{6.6978}\\
		\hline
	\end{tabular}}
	\label{tab:ablation2}
\end{table}

\begin{table*}[h]
	\centering
	\caption{Ablation study of the model structure on the Hong Kong dataset. AP represents Attention Pooling.}
	\setlength{\tabcolsep}{1.3mm}{
		\begin{tabular}{ccc|ccc|cccc}
			\hline

                \multirow{2}{1.5cm}{ \centering \textbf{PCAdaIN}}&\multirow{2}{1cm}{ \centering \textbf{AP}}&\multirow{2}{1cm}{ \centering \textbf{CMoE}}&\multicolumn{3}{c|}{\textbf{Pixel}}&\multicolumn{4}{c}{\textbf{Instance}}\\
                     &&&\textbf{Acc (\%) $\uparrow$}&\centering \textbf{mF1 (\%) $\uparrow$}&\textbf{mIoU (\%) $\uparrow$}&\textbf{Acc (\%) $\uparrow$}&\textbf{mF1 (\%) $\uparrow$}&\textbf{RMSE $\downarrow$}&\textbf{MAE $\downarrow$}\\
                \hline
                 &  & & 62.27 & 60.05 & 43.21 & 55.08 & 53.56 & 10.2712 & 7.1624\\
                 \checkmark &  & & 64.06 & 62.31 & 45.39 & 56.49 & 55.23 & 10.0003 & 6.9430\\
                  & \checkmark &  & 63.89 & 62.11 & 45.23 & 56.35 &54.94 & 9.9073 & 6.9147\\
                  &  & \checkmark & 62.53 & 61.18 & 44.16 &  55.96 & 55.08 & 10.1195 & 7.0564\\
                 \checkmark & \checkmark& & 64.34 & 62.69 & 45.78 & 56.48 & 55.30 & 9.9104 & 6.8473\\
                 \checkmark &  \checkmark &  \checkmark & \textbf{65.52} & \textbf{63.63} & \textbf{46.80} & \textbf{58.37} & \textbf{57.07} & \textbf{9.6919} & \textbf{6.6978}\\
		\hline
	\end{tabular}}
	\label{tab:ablation3}
\end{table*}

Table~\ref{tab:ablation3} presents the ablation study of the proposed model structure by evaluating the effectiveness of three key components, namely PCAdaIN, Attention Pooling, and CMoE. Introducing each module consistently improves the overall performance, demonstrating their effectiveness in \jl{multi-source data} feature fusion and instance-level feature aggregation. Specifically, PCAdaIN, Attention Pooling, and CMoE improve pixel-level Accuracy by approximately 1.8\%, 1.6\%, and 0.3\%, and instance-level Accuracy by 1.4\%, 1.3\%, and 0.9\%, respectively. \jl{Although CMoE achieves smaller Accuracy improvements than the other modules, this is expected since satellite imagery already provides strong visual representations and auxiliary modalities mainly offer complementary information. The consistent improvements across multiple evaluation metrics demonstrate the effectiveness of CMoE in fusing heterogeneous multi-source data features.} We further observe that the design of the instance-level estimation branch also affects pixel-level prediction performance, which indicates that the proposed model is optimized as an integrated framework through end-to-end joint training. In addition, combining PCAdaIN and Attention Pooling yields more gains than using either module individually, highlighting their complementary effects. Finally, integrating CMoE on top of these two components further improves pixel-level Accuracy and instance-level Accuracy by 1.2\% and 1.9\%, respectively. These results verify that each module contributes positively to the framework, while their combined integration provides the most effective solution for both pixel-level and instance-level building age estimation.

\subsection{Computational Complexity Analysis}

\begin{table}[t]
	\centering
	\caption{Computational complexity of the compared methods.}
	\setlength{\tabcolsep}{1.5mm}{
		\begin{tabular}{cccccc}
			\hline
			\multirow{2}{1cm}{\textbf{Models}}&\textbf{Complexity $\downarrow$}&\textbf{Memory $\downarrow$}&\textbf{Parameters $\downarrow$}&\textbf{Speed $\uparrow$}\\
			&\textbf{(GFLOPs)}&\textbf{(MB)}&\textbf{(M)}&\textbf{(FPS)} \\
			\hline
			Unet	&6551.17	&6441.47&	39.42 &	7.780\\
            Unetformer	&238.03	&499.17	&11.72	&76.613\\
            ABCNet	&312.52&	507.22	&13.39	&86.644\\
            Segformer	&1135.27	&5079.99	&27.35	&11.922\\
            CMNeXt	&1162.14	&3625.27&	58.67&	9.249\\
            TransUnet&	2588.87&	2104.17	&105.91	&6.874\\
            CMTFNet&	661.39&	2344.99&	30.07	&17.611\\
            CMT	&598.34	&801.24&	27.68&	19.912\\
            Deeplab	&961.38	&760.70&	39.05&	34.305\\
            A2FPN	&836.65&	1754.69	&12.15	&69.534\\
            CMX&	1140.71	&3660.51&	66.56&	8.342\\
            Asymformer 	&612.26	&1888.87&	33.03	&21.908\\
            MFNet	&2156.13	&2930.55	&18.62	&3.026\\ 
			\hline
            Ours&	6036.38	& 4945.82 & 34.12&	6.126\\

			\hline
	\end{tabular}}
	\label{tab:complex}
\end{table}

We conduct a thorough comparison of computational complexity among the models, in terms of model complexity, memory usage, parameter count, and processing speed. For the parameter comparison, only the trainable parameters of each model are counted, and the detailed results are presented in Table~\ref{tab:complex}. Primarily due to the adoption of a foundation model, our method exhibits relatively high computational complexity and memory consumption. However, its parameter count and inference speed remain comparable to those of existing methods. Overall, the results indicate that our model achieves strong accuracy without excessive computational cost, making it suitable for city-scale data processing.

\section{Discussion}\label{sec:discussion}

\subsection{Implications}

Our framework formulates building age estimation as a segmentation-based classification task, with the predicted results represented as period-based age categories. This representation is consistent with practical applications, as buildings constructed in similar periods often share comparable construction standards, materials, and energy-related characteristics. For example, construction-period information is widely used in building energy assessment, where buildings from different periods are assigned distinct energy performance characteristics for estimation and analysis. Beyond its relevance to building energy assessment, building age information can also contribute to environmental sustainability by helping identify older building stocks that may have lower energy efficiency, supporting the assessment of potential energy-saving and carbon-reduction opportunities, and informing strategies for sustainable urban renewal. In addition, the generated age categories can assist urban renewal by identifying clusters of older buildings, support maintenance prioritization by locating potentially aging building stocks, and facilitate city-scale urban development analysis. Furthermore, the age categories effectively reflect the actual construction ages of buildings, as demonstrated by the favorable RMSE and MAE values achieved in our evaluation. These results demonstrate that the proposed framework provides valuable period-based classification results while also offering reliable estimates of specific building construction ages, with potential applications in urban planning, environmental assessment, and sustainable urban development.

\subsection{Limitations and future work}
Our model achieves superior performance in addressing the problem of city-level building age analysis. However, several challenges remain and warrant further investigation in future work.
\subsubsection{Domain adaptation}
Currently, our age estimation approach primarily relies on the stylistic evolution of buildings and their surrounding environmental context during urban development. However, firstly, both architectural styles and the surrounding environmental context of buildings can vary significantly across geographically distant regions, resulting in substantial visual diversity and making cross-regional generalization more challenging. Secondly, the age distribution of buildings differs among regions, implying that the definition of age classes should be adapted according to the local construction history. These factors make it difficult to directly apply a single model across different regions or to perform effective domain adaptation. Future research should therefore focus on improving cross-city generalization and developing domain adaptation strategies for large-scale building age estimation.

\subsubsection{Model structure}
In the proposed framework, raster data at the pixel level and property data at the instance level are jointly utilized for building age estimation. However, our framework remains limited to these modalities and has not yet explored broader heterogeneous data sources. Future work will focus on extending the framework with flexible interfaces for incorporating additional heterogeneous inputs, such as richer visual, geometric, or semantic information. Moreover, more effective cross-modal interaction mechanisms can be developed within the model to better exploit complementary relationships among modalities, thereby further improving the robustness and accuracy of building age prediction.

\section{Conclusion} \label{sec:conclusion}
In this paper, \jl{we propose a segmentation-based framework that models the relationship between building external characteristics and age, enabling multi-building age estimation within a unified pipeline. The framework utilizes widely available multi-source geospatial data, including remote sensing data and their derived products, as well as building footprint information.} Pixel-level age estimation enables prediction without individual footprints, while its supervision provides auxiliary guidance for instance-level generation, jointly improving building age estimation performance. Specifically, a large pretrained model with a fine-tuning adapter is adopted as the backbone for feature extraction, followed by raster-derived data fusion and property-derived data fusion modules to support pixel-level and instance-level estimations, respectively. 
We employ a pyramid decoder with CMoE modules to integrate \jl{multi-source data} for pixel-level estimations, capturing fine-grained differences between buildings of different ages. In instance-level estimation, we use attention pooling to extract instance-level features and fuse them with attribute features via PCAdaIN, thereby emphasizing the intrinsic characteristics of buildings while reducing interference from the surrounding environment.
Finally, we use TOCL to reduce the ordinal discrepancy between the predicted building age and the ground truth. To comprehensively evaluate the proposed framework, we further construct a building age benchmark using \jl{multi-source geospatial data} in Hong Kong and subsequently perform age estimation for all buildings across the city. Experimental results on both the open-source Amsterdam dataset and the proposed labeled Hong Kong dataset demonstrate the superior performance of the proposed approach.

\section*{Acknowledgements}
This study was jointly supported by the National Natural Science Foundation of China (W2412136, U25A20802) and the Research Grants Council (RGC) of Hong Kong, China (HKU17613022).

\section*{Conflicts of Interest}
The authors declare that there is no conflict of interest regarding the publication of this article.

\printcredits

\bibliographystyle{cas-model2-names}

\bibliography{references}

\end{document}